\documentclass{article}

\usepackage[numbers, sort&compress]{natbib}

\usepackage[final]{neurips_2024}

\usepackage[utf8]{inputenc} 
\usepackage[T1]{fontenc}    
\usepackage{url}            
\usepackage{booktabs}       
\usepackage{amsfonts}       
\usepackage{nicefrac}       
\usepackage{microtype}      

\usepackage{multirow}
\usepackage{adjustbox}
\usepackage{graphicx}
\usepackage{array}
\usepackage{amsmath}

\usepackage{tabularx}
\usepackage{xspace}
\usepackage{xcolor}         
\definecolor{citecolor}{rgb}{0.21,0.49,0.74}
\definecolor{linkcolor}{rgb}{0.8,0.16,0.16}

\usepackage[pagebackref=true,breaklinks=true,colorlinks, citecolor=citecolor,linkcolor=linkcolor,bookmarks=false]{hyperref}

\makeatletter
\DeclareRobustCommand\onedot{\futurelet\@let@token\@onedot}
\def\@onedot{\ifx\@let@token.\else.\null\fi\xspace}

\def\eg{\emph{e.g}\onedot} 
\def\ie{\emph{i.e}\onedot} 
 
\def\etc{\emph{etc}\onedot} 
\def\wrt{w.r.t\onedot} 

\makeatother

\usepackage[capitalize]{cleveref}
    \crefname{section}{Sec.}{Secs.}
    \Crefname{section}{Section}{Sections}
    \Crefname{table}{Table}{Tables}
    \crefname{table}{Table}{Tables} 

\makeatletter
\def\NAT@spacechar{~}
\makeatother

\usepackage{enumitem}
\setlist[itemize]{leftmargin=5mm, itemsep=0.0mm}

\usepackage[nottoc]{tocbibind}
\usepackage{minitoc}
\usepackage{bbm}

\usepackage{pifont}
\usepackage{float}
\newcommand{\cmark}{\ding{51}}%
\newcommand{\xmark}{\ding{55}}%
\usepackage{blindtext}
\usepackage{multicol}
\usepackage{caption}
\usepackage{subcaption}

\usepackage{amsmath,amsfonts,bm}

\def\eqref#1{equation~\ref{#1}}

\def\1{\bm{1}}

\DeclareMathAlphabet{\mathsfit}{\encodingdefault}{\sfdefault}{m}{sl}
\SetMathAlphabet{\mathsfit}{bold}{\encodingdefault}{\sfdefault}{bx}{n}

\definecolor{DarkGreen}{rgb}{0.55, 0.71, 0.0}
\definecolor{DarkRed}{rgb}{0.82, 0.1, 0.26}
\newcommand{\ourbenchmark}{{\textbf{Our benchmark}}}

\newcommand{\ours}{\textsc{ChopinLLM}\xspace} 

\usepackage[nottoc]{tocbibind}
\usepackage{minitoc}
\usepackage{bbm}

\title{On Pre-training of Multimodal Language Models \\Customized for Chart Understanding}

\author{%
  Wan-Cyuan Fan$^{1,3}$ \ \ \  Yen-Chun Chen$^{2}$ \ \ \  Mengchen Liu$^{2}$ \ \ \  Lu Yuan$^{2}$ \ \ \ 
  Leonid Sigal$^{1,3,4}$ \\
  \\
  $^{1}$UBC \qquad $^{2}$Microsoft \qquad 
  $^{3}$Vector Institute for AI \qquad
  $^{4}$CIFAR AI Chair \\
  \\
  \texttt{\{wancyuan, lsigal\}@cs.ubc.ca} \\
}

\begin{document}

\doparttoc 
\faketableofcontents 

\maketitle

\begin{abstract}
    Recent studies customizing Multimodal Large Language Models~(MLLMs) for domain-specific tasks have yielded promising results, especially in the field of scientific chart comprehension.
These studies generally utilize visual instruction tuning with specialized datasets to enhance question and answer~(QA) accuracy within the chart domain.
However, they often neglect the fundamental discrepancy between natural image-caption pre-training data and digital chart image-QA data, particularly in the models' capacity to extract underlying numeric values from charts.
This paper tackles this oversight by exploring the training processes necessary to improve MLLMs' comprehension of charts.
We present three key findings: (1)~Incorporating raw data values in alignment pre-training markedly improves comprehension of chart data.
(2)~Replacing images with their textual representation randomly, during end-to-end fine-tuning, transfers the language reasoning to chart interpretation skills.
(3)~Requiring the model to first extract the underlying chart data and then answer the question in the fine-tuning can further improve the accuracy.
Consequently, we introduce \ours, an MLLM tailored for in-depth chart comprehension.
\ours effectively interprets various types of charts, including unannotated ones, while maintaining robust reasoning abilities.
Furthermore, we establish a new benchmark to evaluate MLLMs' understanding of different chart types across various comprehension levels.
Experimental results show that \ours exhibits strong performance in understanding both annotated and unannotated charts across a wide range of types.
\end{abstract}

\section{Introduction}

In today's data-driven world, visualizations like bar and pie charts are crucial for deciphering complex datasets.
However, the increasing diversity and complexity of these charts highlights the need for advanced tools to enhance human capabilities in data analysis.
Artificial Intelligence~(AI), particularly Multimodal Large Language Models~(MLLMs)~\citep{liu2024visual,zhang2023internlm, lu2024deepseek, mckinzie2024mm1, zhang2023video, chen2023videollm, team2024chameleon, patil2023gorilla, patil2024goex, functionary, zeng2023large, brohan2023rt}, is increasingly used to automate the understanding of scientific charts, promising more efficient and accurate analysis.
Robust benchmarks are also essential, setting standards and metrics that drive the development and evaluation of these AI tools.

Prior studies have introduced end-to-end neural models aimed at enhancing chart comprehension~\citep{lee2023pix2struct,liu2022matcha, zhou2023enhanced}, such as masked table prediction~\citep{zhou2023enhanced, liu2022deplot}, chart question answering~\citep{masry2023unichart}, and chart de-rendering~\citep{liu2022matcha}.
These models specialize in handling one task each within the domain of chart analysis.
Furthermore, advancements in Multimodal Large Language Models~(MLLMs), exemplified by LLaVA~\citep{liu2024visual, liu2023improved} and miniGPT~\citep{zhu2023minigpt}, have showcased their versatility in vision-language tasks.
These generalist models undergo a two-stage training process: initially learning visual-language alignment through image-caption pairs, followed by end-to-end fine-tuning using image-QA pairs.
This training not only enables LLMs to interpret visual data but also retains their extensive pre-trained knowledge, which supports their reasoning abilities and leads to strong performance across diverse visual language understanding tasks.

\begin{figure*}[t]
  \centering
  \includegraphics[page=1, trim={0 5 0 5}, clip, width=0.9\textwidth]{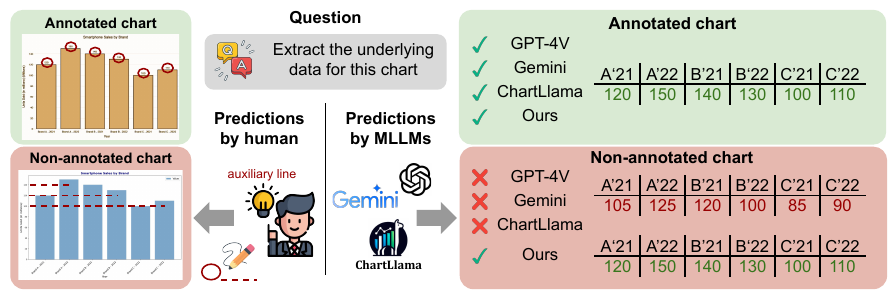}
  \vspace{-1mm}
  \caption{The underlying data values can be inferred regardless of whether the chart is annotated. However, existing MLLMs rely on annotations and struggle with unannotated charts. In contrast, our model bridges this fundamental discrepancy between natural image-caption pre-training data and digital chart image-QA data, enabling it to extract values regardless of whether the chart is annotated.
  } 
  \label{fig:teaser}
  \vspace{-5mm}
\end{figure*}

Recent advancements have further ignited interest in tailoring MLLMs to specialized domains such as scientific chart understanding.
\citet{han2023chartllama,liu2023mmc} have explored collecting instruction-tuned chart data and low-rank adaptation~\citep{hu2021lora} to enhance MLLMs' proficiency with unique chart characteristics. However, research on the fundamental-training regimes -- namely, pre-training to align across modalities and comprehensive end-to-end fine-tuning -- for chart-specific understanding remains scarce.
As shown in~\cref{fig:teaser}, existing MLLMs often struggle to extract the underlying data from charts when numerical values are not annotated.
We hypothesize that this issue stems from a gap in vision-language alignment between natural image-caption pairs and digital chart-data pairs.
Without targeted pre-training for chart-data alignment, models may resort to relying on a ``shortcut'' of recognizing numeric annotations through OCR during fine-tuning with QA pairs, rather than truly understanding the visual subtleties of diverse charts.

This paper addresses the above issues by concentrating on the essential training methodologies for MLLMs, including cross-modal feature alignment pre-training and comprehensive end-to-end fine-tuning.
Our research is guided by the question, \emph{``How does fundamental MLLM training influence the enhancement of general MLLMs with chart-specific domain understanding?''}
Our findings indicate that: 
(1)~Raw data extraction are pivotal in alignment pre-training to bolster chart data comprehension;
(2)~Substituting some chart images with purely textual data during end-to-end fine-tuning not only preserves LLM's text-only reasoning ability but also augments chart interpretation capabilities;
(3)~Augmenting QAs with data extraction tasks in the fine-tuning phase allows model to achieve the data prompting during testing, where it first extract data and then answer the QAs, further improving the its reasoning skills.

In summary, our key contributions in this paper are: (1) We introduce \ours,\footnote{\textbf{Ch}art \textbf{O}riented \textbf{P}retraining \textbf{In}tegration in \textbf{L}arge \textbf{L}anguage \textbf{M}odels} a Multimodal Large Language Model tailored for comprehensive chart understanding. This model excels at interpreting various chart types including unannotated ones, underpinned by our detailed analysis and training guidance that emphasizes the importance of foundational training for chart-specific tasks. (2) We propose a novel data generation pipeline using text-only Large Language Models to efficiently produce large-scale pairwise data. This approach significantly reduces the costs and complexity of data generation for MLLM training. Furthermore, (3) we establish a robust benchmark comprising a diverse array of chart types and question-answering levels, designed to rigorously evaluate MLLMs' fundamental understanding of the scientific chart domain. 

    




\vspace{-0.05in}
\section{Generating data for chart understanding}

\begin{figure*}[t]
  \centering
  \includegraphics[page=1, trim={35 85 25 80}, clip, width=1.\textwidth]{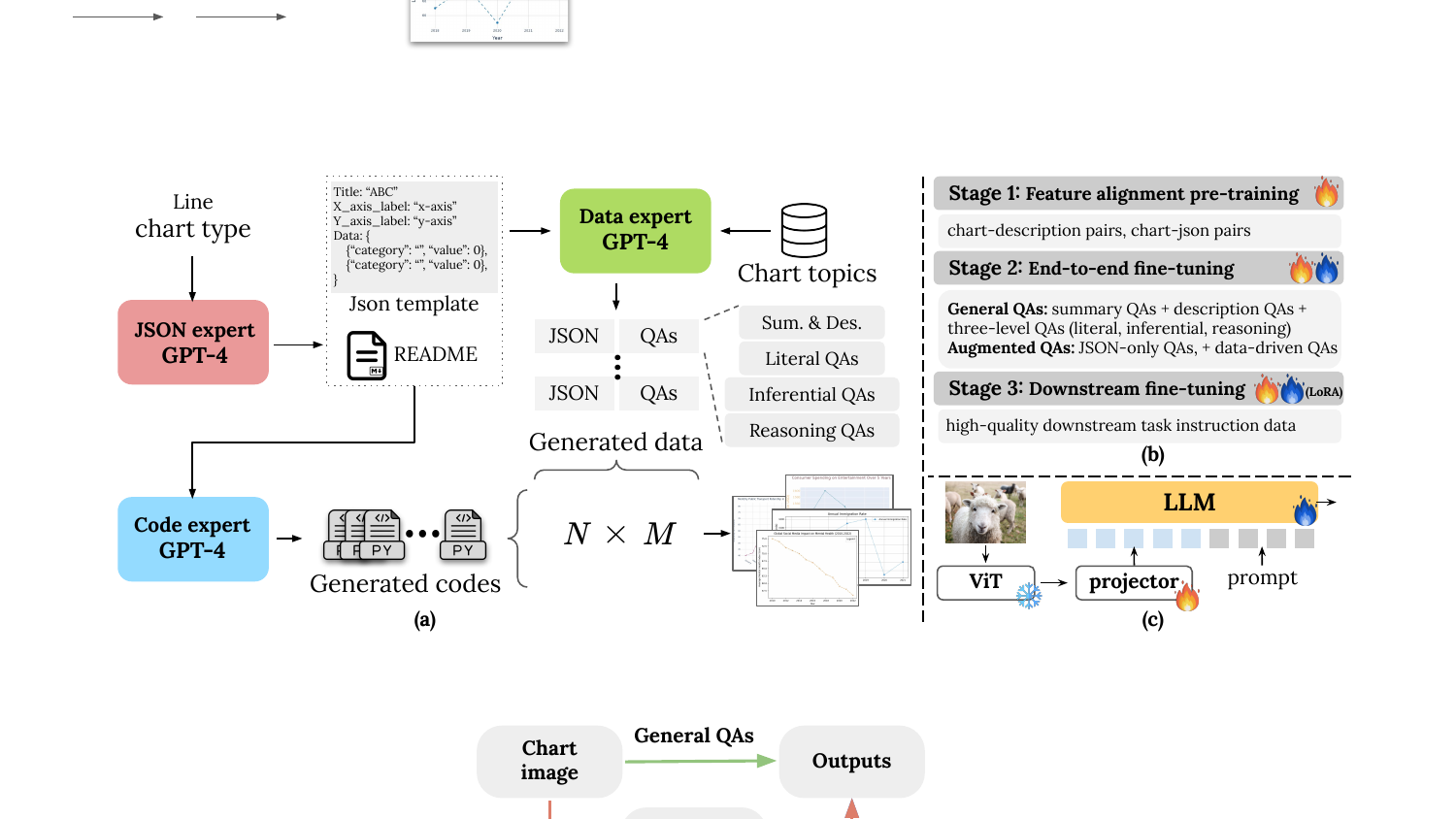}
  \vspace{-7mm}
  \caption{\textbf{Overview of (a) the proposed data generation pipeline and (b) training strategies of \ours}.
  Generating code and data points conforming to a shared JSON template enables quadratic scaling of the data size~(\wrt to \#GPT calls).
  The 3-stage training equips our model to grasp the underlying data, thereby achieving a fundamental understanding of charts.
  ($N$ and $M$ denote the number of generated scripts and data, respectively.)
  } 
  \label{fig:framework}
  \vspace{-3mm}
\end{figure*}

To build a chart understanding MLLM and study its fundamental training process, a comprehensive dataset containing chart images paired with captions and raw data is essential for pre-training, alongside different types of question-answer pairs for end-to-end fine-tuning. However, no existing dataset provides the necessary variety of chart types, topics, and styles. To bridge this gap, we introduce a novel data generation pipeline for large-scale chart data generation~(\cref{method:data}) and QAs generation~(\cref{method:qa}). With the data at hand, we then explore various training strategies in the later sections, including feature alignment pre-training and end-to-end fine-tuning for LLMs.
\Cref{fig:framework} presents an overview of our framework.

\subsection{Efficient data generation with quadratic scaling}
\label{method:data} 

Our data generation leverages the promising text content generation and coding abilities of large language models, \eg, GPT-4, to generate chart images and data. Specifically, LLMs allow us to synthesize raw data for chart images, and then the generated Python script turns the raw data into a chart image. In this way, we can produce image data without accessing costly multimodal LLMs like GPT-4V. Unlike previous and concurrent works~\citep{han2023chartllama, xia2024chartx} that prompt LLMs to iteratively generate CSV data, QAs, and Python script for each chart image -- a process that is costly to massively scale -- our pipeline features parallel code and data generation through shared templates and READMEs for consistent definitions and formats across the same chart types. Most importantly, since all code script and data share the same structure, our generated data can be universally applied to any generated code and vice versa, significantly enhancing scalability without exhaustively prompting LLMs. Please refer to~\cref{sec:append:filtering} for more details about shared templates and the raw data generation.

\subsection{Diverse QA synthesis}
\label{method:qa} 

Based on the parallel data generation pipeline, we are able to collect massive amount of chart image and JSON raw data pairs for the feature alignment pre-training.
Now, we details how we generate different types of QAs for end-to-end fine-tuning.
Specifically, having each JSON data as input, we use text-only LLM to generate question-answer~(QA) pairs.
To cover various question-anwser for chart data, we include general QAs, containing not only description and summary QA but also three different level of QAs: literal QAs, inferential QAs, and reasoning QAs~(as illustrated in~\cref{fig:demo}).
Furthermore, to enhance the training of chart understanding, we introduce two additional augmented QAs~(for training only): text-only QAs and data-driven QAs.
Please refer to~\cref{sec:append:filtering} for the details of each QA type.

\begin{table*}[t]
\scriptsize
\centering
\caption{Comparative analysis with existing benchmarks for chart understanding evaluations. *~denotes unbounded chart types. Chart variation refers to whether the dataset contains chart images with different styles but sharing the same raw data.}
\vspace{-0.06in}
\newcolumntype{C}{>{\centering\arraybackslash}X}
\newcolumntype{L}{>{\raggedright\arraybackslash}X}
\newcolumntype{R}{>{\raggedleft\arraybackslash}X}
\begin{tabularx}{0.98\textwidth}{L r r r c c c}
\toprule
\multirow{2}{*}{Benchmark} & \multirow{2}{*}{\# Image} & \multirow{2}{*}{\# Chart type} & \multirow{2}{*}{\begin{tabular}[c]{@{}c@{}}Avg. \# QAs \\ per image\end{tabular}} & \multirow{2}{*}{\begin{tabular}[c]{@{}c@{}}Multi-level QAs \\ per image \end{tabular}} & \multirow{2}{*}{\begin{tabular}[c]{@{}c@{}}Raw data\\ per image\end{tabular}} & \multirow{2}{*}{\begin{tabular}[c]{@{}c@{}}Chart\\ Variation\end{tabular}} \\ 
 &  &  &  &  &  &  \\
\midrule
PlotQA~\citep{methani2020plotqa} & 33.7k & 3 & 1 & \xmark & \xmark & \xmark \\
ChartQA~\citep{masry2022chartqa} & 1.5k & 3 & 1 & \xmark & \cmark & \xmark \\
Chart-to-text~\citep{kantharaj2022chart} & 6.6k & 6 & 1 & \xmark & \xmark & \xmark \\
MMC~\citep{liu2023mmc} & 2k & 6 & 1 & \xmark & \cmark & \xmark \\
Chartbench~\citep{xu2023chartbench} & 2.1k & 9 & 9 & \cmark & \xmark & \xmark \\
ChartX~\citep{xia2024chartx} & 6k & 18 & 1 & \xmark & \cmark & \xmark \\
CharXiv~\citep{wang2024charxiv} & 2.3k & * & 5 & \cmark & \xmark & \xmark \\
\cmidrule(lr){1-7}
Ours & 5.48k & 20 & 13.5 & \cmark & \cmark & \cmark \\ 
\bottomrule
\end{tabularx}
\vspace{-0.2in}
\label{table:benchmark_comparison}
\end{table*}

\subsection{A new benchmark for comprehensive chart understanding}
\label{sec:benchmark}
A chart expert model should be capable of understanding a wide range of common chart types and, like a human, should not only be able to answer questions of varying complexity but also grasp the underlying data. However, as shown in~\cref{table:benchmark_comparison}, existing chart benchmarks either cover only a limited range of chart types~(\eg, line, bar, and pie charts) or lack comprehensive QA sets to evaluate a model's understanding of charts from various perspectives, including raw data comprehension, inferential abilities, and mathematical reasoning capabilities.
To bridge this gap, we propose a comprehensive benchmark derived from the aforementioned synthetic dataset.
It covers 20 different chart types, three different levels of QAs (literal, inferential, and reasoning QAs), and provides both long and short answers.
Notably, the chart images in the benchmark are not all annotated, allowing assessment of the model's ability to understand the underlying data of a chart as humans do.
To ensure the quality of the images in the benchmark, we employed human evaluations to filter the data and obtain a high-quality test set.
The evaluations are based on two criteria: \textit{Answerability}: whether the question is answerable given the chart image.
\textit{Correctness}: whether the provided answer is correct. Please refer to~\cref{sec:suppl:benchmark} in the supplementary materials for more details about benchmark statistics, filtering, analysis, etc. Note that these QAs equally cover literal, inferential, and reasoning questions for measuring chart understanding of MLLMs.

\section{Experiments and model analysis}

\subsection{Experimental setup}
\paragraph{Benchmark.}

Our evaluation utilizes four classical benchmarks: (1) ChartQA dataset~\citep{masry2022chartqa}, which includes 1.5k chart images in its test set, divided into human-written and machine-generated questions with 1.2k QA pairs each.
The human-written questions often require mathematical reasoning.
(2) ChartQA also provides CSV data for each image, enabling us to conduct a Chart-to-Table task to assess the ability of MLLMs to extract raw data from charts, following previous studies~\citep{han2023chartllama, liu2022deplot}.
(3) Additionally, we use the PlotQA dataset~\citep{methani2020plotqa} where images generally lack numerical value annotations, necessitating value inference relative to the Y-axis.
(4) Lastly, for evaluating the models' capability to capture global concepts, we assess on the Chart-to-Text task using the \textit{Pew} and \textit{Statista} splits from the dataset~\citep{kantharaj2022chart}.
For the evaluation metrics, we follow previous works~\citep{masry2022chartqa, han2023chartllama, kantharaj2022chart} and use relaxed accuracy for QA tasks, as well as the \textit{F1} score and \textit{Relative Number Set Similarity}~(RNSS) for chart-to-table tasks. Additionally, \textit{BLEU-4} is used for the chart-to-text task.



\vspace{-0.1in}
\paragraph{The 3-stage Training Process.}

Unlike previous approaches that convert a general MLLM into a chart-specific expert by only applying LoRA fine-tuning on limited high-quality data~\citep{han2023chartllama}, training \ours unfolds in three stages, illustrated in \cref{fig:framework}~(b).
The 3-stage training enables our model not only to understand chart QAs and downstream tasks but also to capture the underlying data, thereby achieving a fundamental understanding of charts.
In the initial pre-training stage, we fix the ViT and LLM while training the projector from scratch using original LLaVA data alongside our newly generated chart-description and chart-json pairs.
The second stage involves freezing ViT and jointly fine-tuning the projector and LLM with both original LLaVA QA pairs and our generated chart QA pairs, enabling the LLM to comprehend visual tokens and facilitate chart question answering.
Finally, we apply LoRA fine-tuning to align the LLM’s response distribution with the target downstream dataset. 
In this section, we showcase the model's performance at the final stage on the downstream tasks. For the results of stages 1 and 2, please refer to~\cref{method:pretraining} and~\cref{method:finetuning}.

\begin{table}[t]
\scriptsize
\centering
\caption{\textbf{Comprehensive evaluation across four chart benchmarks.}
H and A denote the human and augmented branch in ChartQA, respectively.
Stat. represent the statista split.
$^\dagger$: our reproduction using the official code. Note that for fair comparison, we don't use chain-of-reasoning in the inference.
The best result is highlighted in \textbf{Bold} and the second \underline{underlined}. \# chart data denotes the number of pairwise chart data used in the training. A and S in the data source represent annotated data and synthetic data, respectively. 
}
\vspace{0mm}
\newcolumntype{C}[1]{>{\centering\arraybackslash}p{#1}}
\newcolumntype{L}[1]{>{\arraybackslash}p{#1}}
\newcolumntype{R}{>{\raggedleft\arraybackslash}X}
\begin{tabularx}{1.0\columnwidth}{p{3.2cm} L{3mm}c ccc cc cc C{4mm}C{4mm}}
\toprule
\multirow{2}{*}{Method} & \multirow{2}{*}{\begin{tabular}[c]{@{}c@{}} Data \\ \# \end {tabular}} & \multirow{2}{*}{\begin{tabular}[c]{@{}c@{}} Data \\ Source\end {tabular}} & \multicolumn{3}{c}{ChartQA} & \multicolumn{2}{c}{Chart-to-Table} & \multicolumn{2}{c}{Chart-to-Text}  & \multicolumn{2}{c}{PlotQA$^*$} \\
\cmidrule(lr){4-6}  \cmidrule(lr){7-8}  \cmidrule(lr){9-10} \cmidrule(lr){11-12}  
 & & & H & A & Avg. & F1 & RNSS & Pew & Stat. & v1 & v2 \\
 \midrule
Pix2struct~{\tiny \citep{lee2023pix2struct}} & 80M & A & 30.50 & 81.60 & 56.00 & - & - & 10.30 & 38.00 & - & - \\
Matcha~{\tiny \citep{liu2022matcha}} & 16M & S+A & 38.20 & \underline{90.20} & 64.20 & - & - & 12.20 & 39.40 & - & - \\
Unichart~{\tiny \citep{masry2023unichart}} & 7M & S+A & 43.92 & 88.56 & 66.24 & 52.71 & - & 12.48 & 38.21 & - & - \\ 
DePlot~{\tiny \citep{liu2022deplot}} & 0.5M & S+A & - & - & - & 87.22 & 94.28 & - & - & - & - \\ \midrule
LLaVA$_{\text{7B}}$$^\dagger${\tiny \citep{liu2024visual}} & - & - & 36.00 & 67.44 & 51.72 & 56.96 & 91.83 & 8.50 & 21.50 & 27.26 & 30.64 \\
LLaVA$_{\text{13B}}$ & - & - & 37.68 & 72.96 & 55.32 & 48.95 & - & 7.16 & 24.65 & - & - \\
LLaVA$_{\text{13B}}$$^\dagger$ & - & - & 42.56 & 73.60 & 58.08 & 63.18 & 93.18 & 8.83 & 22.39 & 27.68 & 30.98 \\
ChartLlama$_{\text{13B}}$~{\tiny \citep{han2023chartllama}} & 0.16M & A & 48.96 & \underline{90.36} & 69.66 & \underline{89.84} & \underline{94.65} & \underline{14.23} & 40.71 & 29.76 & 29.93  \\
MMC$_{\text{7B}}$~{\tiny \citep{liu2023mmc}} & 0.6M & S+A & - & - & 57.40 & - & - & - & - & - & -  \\
ChartInstruct$_{\text{7B}}${\tiny\citep{masry2024chartinstruct}} & 0.19M & A & 45.52 & 87.76 & 66.64 & 18.87 & 34.59 & 13.83 & \textbf{43.53} & - & -  \\
ChartAst$_{\text{13B}}$~{\tiny \citep{meng2024chartassisstant}} & 24M & S+A & \textbf{65.9} & \textbf{93.9} & \textbf{79.9} & \textbf{91.6} & - & \textbf{15.5} & \underline{41.0} & - & - \\
\midrule
\ours$_{\text{7B}}$ & 5M & S & 52.28 & 87.68 & 69.98 & 83.63 & \underline{95.27} & 11.50 & 38.97 & \underline{30.06} & \underline{31.08} \\
\ours$_{\text{13B}}$ & 5M & S & \underline{54.11} & 88.67 & \underline{71.39} & 88.12 & \textbf{95.95} & 12.66 & 40.81 & \textbf{33.98} & \textbf{33.96} \\
\bottomrule
\end{tabularx} 
\vspace{-5mm}
\label{exp:main} 
\end{table}

\subsection{Downstream fine-tuning}

We build \ours with the best setting based on the observation in the stage 1 and 2~(the data used in each stage can be found in~\cref{fig:framework}~(b)), and we compare \ours with existing chart understanding approaches, including Pix2struct~\citep{lee2023pix2struct}, Matcha~\citep{liu2022matcha}, Unichart~\citep{masry2023unichart}, Deplot~\citep{liu2022deplot}, LLaVA~\citep{liu2024visual}, and ChartLlama~\citep{han2023chartllama}. The results are shown in~\cref{exp:main}. 

\vspace{-0.1in}
\paragraph{Classical question-answering on ChartQA.}
We find that \ours achieves the second best performance on ChartQA, as shown in~\cref{exp:main}. Notably, compared to the recent work of ChartAst, we use significantly less data, and most importantly, our training data is fully synthetic, requiring no additional human effort. In comparison to the third-best model, ChartLlama, we outperform it by $\approx 5\%$ on the human split of ChartQA. Note that the human split in ChartQA is more challenging than the augmented split, as it contains more reasoning questions, suggesting that \ours is better at performing reasoning tasks.

\vspace{-0.1in}
\paragraph{Raw data and global concept understanding.}
As shown in~\cref{exp:main}, \ours achieves the competitive F1 score and the highest RNSS result, indicating that \ours can capture not only the structure but numerical values of raw data of chart images.
We note that the performance on the chart-to-table task may have been saturated, as the images are mostly annotated.
In this context, this primarily measures the OCR capability and does not assess the ability to capture the underlying data.
As for the Chart-to-Text, shown in~\cref{exp:main}, \ours performs comparable in the global concept capturing and can caption chart image with meaningful texts.

\vspace{-0.1in}
\paragraph{Performance on unannotated chart images.}
Most of the images in ChartQA~\citep{masry2022chartqa} are annotated, which means the numerical values of data points are explicitly shown on the images.
We observe that existing chart MLLMs, such as ChartLlama~\citep{han2023chartllama}, seem to heavily rely on this annotation for chart understanding, which is not ideal since real-world charts may be unannotated.
We further evaluate them using the PlotQA dataset, and the results are shown in the last column of~\cref{exp:main}.
Notably, since training previous models like ChartLlama on PlotQA is infeasible, we load the model weights as used in ChartQA and perform zero-shot prediction on PlotQA.
The results show that our model performs significantly better~($\approx3\%$ improvement) on unannotated chart images, suggesting that our methods with fundamental training rely less on numerical annotations. Note that the comparison with ChartAst and ChartInstruct is not included, as it was trained on PlotQA, which would affect the validity of the zero-shot predictions on PlotQA.



\section{Conclusion}
In this paper, we explore the impact of fundamental training strategies in adapting generalist Multimodal Large Language Models~(MLLMs) to chart understanding.
We offer practical guidance for optimizing feature alignment pre-training and end-to-end fine-tuning.
Leveraging these enhanced training strategies, we introduce a specialized chart MLLM, named \ours, capable of interpreting diverse chart types independently of numerical annotations.
Extensive experiments confirm that \ours surpasses the previous state-of-the-art across four benchmarks, validating our framework's effectiveness.
Additionally, we present a new benchmark specifically designed to evaluate MLLMs' comprehension across various chart types and multiple levels of understanding.


\bibliography{sources/iclr2025_conference}
\bibliographystyle{sources/iclr2025_conference}

\appendix

\clearpage

\begingroup
\hypersetup{colorlinks=false, linkcolor=black}
\hypersetup{pdfborder={0 0 0}}
\part{} 
\parttoc 
\endgroup

\setcounter{page}{1}
\def\thesection{\Alph{section}}
\renewcommand{\thetable}{A\arabic{table}}
\renewcommand{\thefigure}{A\arabic{figure}}


\clearpage

\section{Details of the data generation pipeline}
\label{sec:append:filtering}

\subsection{Details of efficient data generation with quadratic scaling}

\paragraph{Shared Template and README.} As shown in~\cref{fig:framework}~(a), given a chart type~(\eg, line) sampled from a predefined chart type database, the JSON expert GPT-4 first generates a JSON template for the given chart type, along with a README file. In detail, the JSON template contains general information for the chart image, including the title, x-axis, y-axis information, and raw data. The README contains the definition of the chart type and the meanings of the keys and values to enhance understanding of the JSON template. Please refer to~\cref{suppl:dataset:template} for some examples. We note that the JSON template, together with the README, ensures the consistency of data generation so that further data and code generation can follow the explicit format and definition guidance of the template data. Note that we choose JSON as our primary data representation format, in contrast to previous works~\citep{han2023chartllama, masry2022chartqa, methani2020plotqa, xia2024chartx}, which used CSV. The JSON format allows us to incorporate not only numerical data but also additional chart information, such as titles and the scales of x and y axes, which is beneficial for pair-wise pre-training tasks. Moreover, JSON data is structured, and when paired with a README file, it minimizes ambiguity in data descriptions, which is particularly valuable for complex chart types. For instance, in candlestick charts, we can clearly define a data point as a dictionary containing ``open'', ``close'', ``high'', and~``low'' values, rather than a list where the meaning of each number might be unclear.

\paragraph{Orthogonal Data and Code Generation.} With the template files at hand, we can generate data and code independently. For the data generation branch, to ensure the generated data covers diverse topics, we jointly input the produced template files~(\ie, JSON template and README) and a topic sampled from a pre-defined topic set (\eg, energy production and market share) into a data expert GPT-4 module. For the complete topic list, please refer to~\cref{suppl:dataset:topics}. We require the data expert GPT-4 to  follow the definitions in the template files and generate $M$ JSON data along with different kinds of questions and answers~(\eg, summary QA) based on the raw data. As for code generation, another code expert GPT-4 is utilized to produce $N$ Python code based on the given chart type, data template, and Python library. Note that to prevent generating simple code repeatedly for the given chart type, we explicitly ask the code expert GPT-4 to introduce visual variations in aspects such as color, legend, grid, font, and mark texture, \etc. More details can be found in the Appendix.

\subsection{Dataset filtering}
We provide the details for each QA type as follows:

\begin{itemize}
    \item \textbf{Description QAs:} Generate objective descriptions based on the chart data.
    \item \textbf{Summary QAs:} Summarize the chart, highlighting key findings.
    \item \textbf{Literal QAs:} Extract specific values directly from the data.
    \item \textbf{Inferential QAs:} Infer global insights, such as identifying extreme values.
    \item \textbf{Reasoning QAs:} Perform calculations to derive answers from chart data.
    \item \textbf{JSON-only QAs:} 
    Replace images with JSON raw data to augmented previous QAs.
    \item \textbf{Data-driven QAs:} 
    Prompt the model to extract JSON raw data before answering the question.
\end{itemize}

These QAs encompass a range of questions for chart images, covering abilities from basic data understanding and global concept comprehension to advanced reasoning, allowing us to further assess the abilities of MLLMs. Note that, for each QA pair, we use GPT-4 to generate both long and short answers. The long answer, generated first, includes a step-by-step explanation to derive the answer, while the short answer, generated later, contains only the final answer derived from the long explanation. Short answers contain only numerical values or Yes/No response for convenient evaluation purpose. For more examples of generated chart and QAs, please refer to~\cref{suppl:dataset:demo}.

\subsection{Dataset filtering}
In~\cref{method:data}, we introduce a novel data generation pipeline that leverages text-only LLMs. This pipeline enables us to collect chart images along with various data and QA pairs without extensive human effort, thereby reducing the cost of creating pairwise data. However, LLMs are not perfect and can make mistakes in either data generation or code script generation. Thus, in this section, we discuss the data filtering techniques we use to improve the quality of the synthetic dataset. The generation pipeline is split into three parts: shared template generation, data and QA generation, and code script generation. We now detail the filtering process for each part.

\paragraph{Shared template and README}
The shared template and README file for each chart type form the core of the entire data generation process, as the subsequent raw data, QA, and Python script are based on the shared template. Therefore, for the shared template, we deploy a human check to ensure the template contains necessary elements for the chart (i.e., title, x-axis, y-axis, data). Additionally, humans are required to verify the correctness of the chart type definitions in the README. Note that we consider 20 different chart types; thus, there are 20 template JSON files along with the READMEs in our dataset.

\begin{figure*}[t]
  \centering
  \includegraphics[page=2, trim={95 75 90 15}, clip, width=0.9\textwidth]{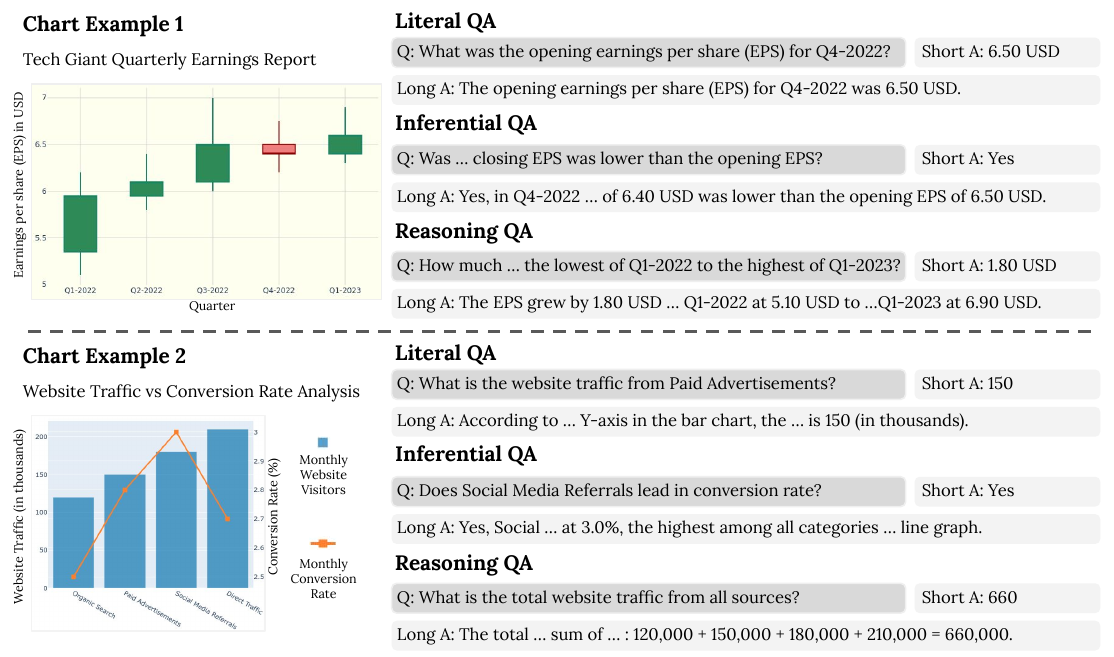}
  \vspace{-3mm}
  \caption{\textbf{Examples of generated three-level QAs with long and short answers}, accessing the understanding of charts from various perspectives. 
  Best viewed in color.}
  \label{fig:demo}
  \vspace{-4mm}
\end{figure*}

\paragraph{Data and QA generation}
Since all the data should follow the template JSON, for the data generation part, we apply filtering based on the JSON structure. Specifically, we remove generated data that deviates from the template file by comparing the elements in the keys of the JSON dictionary and the data types of all the values. As for QA generation, we check the structure of the output dictionary. In detail, the keys of the output QA dictionary should contain summary, description, literal, inferential, and reasoning QAs. We filter out QAs with missing attributes.

\paragraph{Code script generation}
We predefined a code expert GPT-4 to use four different Python libraries to plot the chart images: Matplotlib, Plotly, Pygal, and Seaborn. The advantage of these libraries is that if the Python code or input data is incorrect in terms of structure or other errors, the generated image will either be missing with a Python error or display a "No Data" icon. Thus, we apply a two-step filtering process: (1) Python Error Filtering: If there is an error while running the Python script to generate the image, we will remove the script and the corresponding JSON data. (2) OCR Tool Filtering: If the image is generated but there is some other error, the output image will display a "No Data" icon on it. To this end, we further use an OCR tool to detect whether there is any "No Data" icon in the images. If so, we will remove the data and script accordingly.

\begin{table*}[t]
\scriptsize
\centering
\caption{Comparative analysis with existing benchmarks for chart understanding evaluations. *~denotes unbounded chart types. Chart variation refers to whether the dataset contains chart images with different styles but sharing the same raw data.}
\vspace{-0.06in}
\newcolumntype{C}{>{\centering\arraybackslash}X}
\newcolumntype{L}{>{\raggedright\arraybackslash}X}
\newcolumntype{R}{>{\raggedleft\arraybackslash}X}
\begin{tabularx}{0.98\textwidth}{L r r r c c c}
\toprule
\multirow{2}{*}{Benchmark} & \multirow{2}{*}{\# Image} & \multirow{2}{*}{\# Chart type} & \multirow{2}{*}{\begin{tabular}[c]{@{}c@{}}Avg. \# QAs \\ per image\end{tabular}} & \multirow{2}{*}{\begin{tabular}[c]{@{}c@{}}Multi-level QAs \\ per image \end{tabular}} & \multirow{2}{*}{\begin{tabular}[c]{@{}c@{}}Raw data\\ per image\end{tabular}} & \multirow{2}{*}{\begin{tabular}[c]{@{}c@{}}Chart\\ Variation\end{tabular}} \\ 
 &  &  &  &  &  &  \\
 \midrule
PlotQA~\citep{methani2020plotqa} & 33.7k & 3 & 1 & \xmark & \xmark & \xmark \\
ChartQA~\citep{masry2022chartqa} & 1.5k & 3 & 1 & \xmark & \cmark & \xmark \\
Chart-to-text~\citep{kantharaj2022chart} & 6.6k & 6 & 1 & \xmark & \xmark & \xmark \\
MMC~\citep{liu2023mmc} & 2k & 6 & 1 & \xmark & \cmark & \xmark \\
Chartbench~\citep{xu2023chartbench} & 2.1k & 9 & 9 & \cmark & \xmark & \xmark \\
ChartX~\citep{xia2024chartx} & 6k & 18 & 1 & \xmark & \cmark & \xmark \\
CharXiv~\citep{wang2024charxiv} & 2.3k & * & 5 & \cmark & \xmark & \xmark \\
\cmidrule(lr){1-7}
Ours & 5.48k & 20 & 13.5 & \cmark & \cmark & \cmark \\ 
\bottomrule
\end{tabularx}
\vspace{-0.08in}
\label{table:suppl:benchmark_comparison}
\end{table*}
\section{Details of the generated benchmark}
\label{sec:suppl:benchmark}
\subsection{Motivations and Design Principle}
To measure the comprehensive chart understanding ability of MLLMs, we assume that models can understand the underlying data of various chart types and perform QAs related to charts. These QAs shouldn't be limited to simple questions about the title or x-axis; instead, they should range from basic to advanced QAs, requiring models to have a global conceptual understanding or even perform mathematical reasoning. However, existing benchmarks either lack a diverse range of chart types or fail to provide comprehensive QAs for each image, lack of a full assessment of understanding from multiple perspectives. To this end, we leverage our powerful data generation pipeline to generate a small set of data and apply filtering through an automatic pipeline and human evaluation. The resulting benchmark has several features to facilitate research in scientific chart understanding: (1) 20 different chart types, covering general use cases to scientific reports; (2) comprehensive QAs for each image, including literal, inferential, and reasoning questions, similar to the design of the training data; and (3) raw data for each image, measuring the models' ability to understand the underlying data. We provide a comparison with previous benchmarks in~\cref{table:suppl:benchmark_comparison}.

\begin{figure*}[t]
  \centering
  \includegraphics[page=5, trim={0 100 0 100}, clip, width=1.\textwidth]{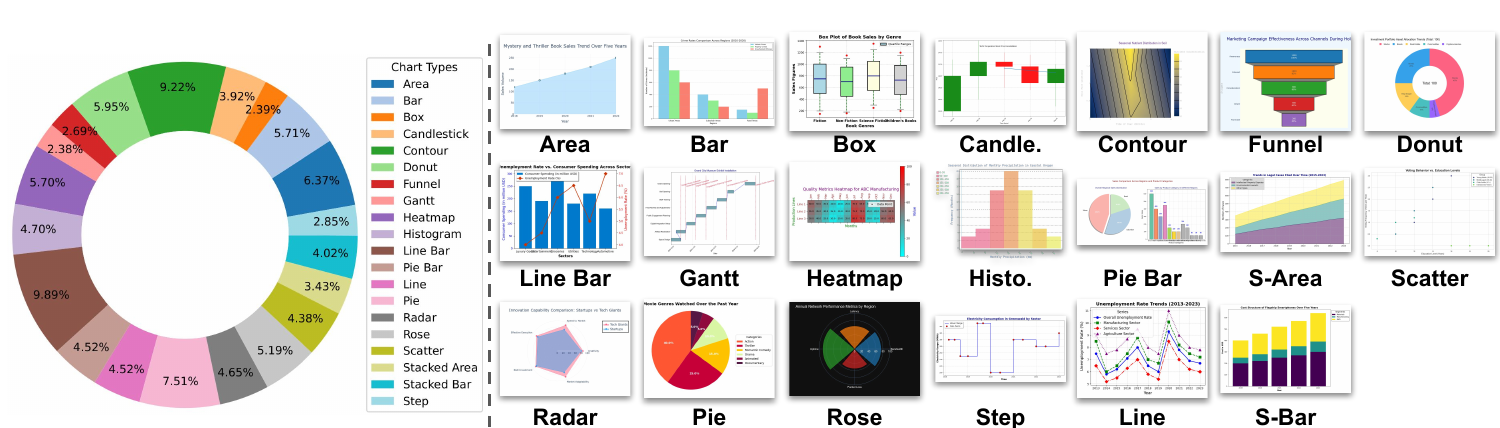}
  \vspace{-5mm}
  \caption{\textbf{Overview of our benchmark}. \textbf{Left}: Data distribution across different chart types. \textbf{Right}: Comprehensive list of chart types in the dataset with examples.}
  \label{fig:append:benchmark}
\end{figure*}

\subsection{Dataset statistics}
We provide the data statistics after filtering in~\cref{table:suppl:benchmark_comparison}. Additionally, a comprehensive list of all the chart types and the data distribution for each type is provided in~\cref{fig:append:benchmark}. Specifically, for each chart type, we initially create $M=30$ JSON data and $N=12$ Python scripts. For each JSON dataset, we generate 15 QAs. As mentioned in~\cref{method:data}, since all code scripts and data share the same structure, our generated data can be universally applied to any generated code and vice versa. In this way, we obtain approximately $360$ unique chart images and around $5.4k$ chart-QA pairs for each chart type. After data filtering, we retain $74k$ chart-QA pairs in total, resulting in an average of $3.5k$ pairs per chart type and an average of $13.5$ QAs per image. By applying multiple Python scripts to the same data point, our benchmark includes 608 unique data points, with an average of $9.2$ variations per data point.

\subsection{Data filtering}
The benchmark is generated by our data generation pipeline, while the LLMs are not faultless, resulting in some noisy question-answer pairs. Unlike training data, noisy data can be problematic for benchmark, and cannot accurately access the model performance. Thus, to make sure the quality of the benchmark, we leverage automatic filtering process and human evaluation to filter chart images and question-answer pairs. We now detail the data filtering process of the proposed benchmark.

\paragraph{General data filtering.} 
In the first stage of data filtering, we leverage the automatic data filtering pipeline used for our training data, as mentioned in~\cref{sec:append:filtering}. Specifically, since the JSON data shares the same structure, we can filter out JSON files with incorrect dictionary structures. Furthermore, since the generated QAs follow a predefined dictionary structure, we can filter out QAs with missing attributes or keys. Lastly, when generating chart images using Python code and JSON, we record any execution errors and further filter out Python scripts that produce error messages.

\paragraph{Human evaluation - image filtering.} 
To make sure the quality of the benchmark, we further adopt human evaluation. Recall that, in our benchmark, we have $608$ unique data points, with an average of $9.2$ variations per data point. To reduce the complexity and cost of human evaluation, we first invite human workers to check the image first, in which human workers have to go through all chart images and check the readibility of the chart. Specifically, workers will have to remove chart images if they have missing data points comparing to other variation or they are draw incorrect (potentially due to incorrect python code).

\paragraph{Human evaluation - QA filtering.} 
After filtering the chart images, we then conduct QA filtering. In this step, we ask human workers to review 608 unique data points, with each data point having 15 QAs, resulting in approximately 9k test pairs. Specifically, for each test pair, we provide the human workers with a chart image along with the corresponding question and answers. Human workers are asked to filter the QA pairs based on two criteria: (1) Answerability: whether the question is answerable given the chart image, and (2) Correctness: whether the provided answer is correct. After collecting the feedback, we perform final data filtering to obtain the benchmark set.

\begin{figure*}[t]
  \centering
  \includegraphics[page=23, trim={0 0 0 0}, clip, width=1.\textwidth]{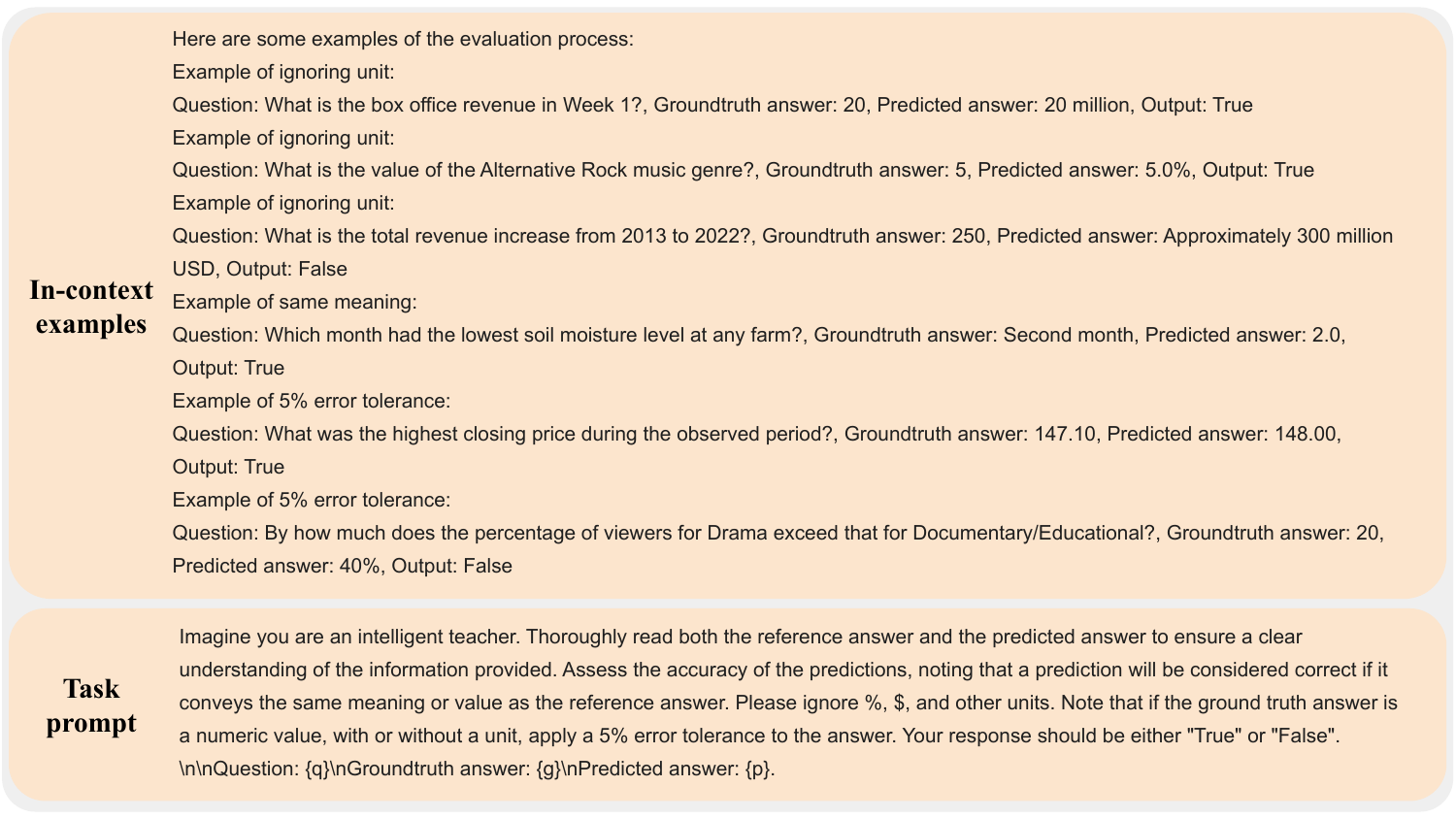}
  \vspace{-6mm}
  \caption{Input prompt example for GPT-Acc. q, g, and p in the task prompt denote the question, ground truth answer, and predicted answer, respectively.} 
  \label{suppl:fig:gpt_acc}
  \vspace{-3mm}
\end{figure*}

\subsection{Evaluation metrics}
To quantitative analyze the performance on our benchmark, we adopt the relaxed accuracy metric for numeric answers,
allowing a $5\%$ margin of error from the exact value, and use exact match for non-numeric answers as per the standard in previous studies. Since the output of the MLLM model can be open form, following previous works~\citep{liu2023mmc, han2023chartllama, xia2024chartx}, we also provide GPT-accuracy (GPT-Acc) using GPT model to further verify the performance. In~\cref{suppl:fig:gpt_acc}, we show how we leverage the LLM to measure the accuracy by providing the detail prompts.\footnote{GPT-4o-mini (2024-07-18)} Lastly, for underlying data understanding task, we follow previous work Deplot~\citep{liu2022deplot} and measure performance using F1 score of Relative Mapping Similarity (RMS) and Relative Number Set Similarity (RNSS) to evaluate numeric accuracy and raw data similarity, respectively. 

\subsection{Performance of Existing Models}

\begin{table}[t]
\scriptsize
\centering
\caption{\textbf{Comprehensive evaluation on our benchmark.} Note that R-Acc and GPT-Acc denote relaxed accuracy and GPT accuracy, respectively.}
\vspace{0mm}
\newcolumntype{C}{>{\centering\arraybackslash}X}
\newcolumntype{L}{>{\raggedright\arraybackslash}X}
\newcolumntype{R}{>{\raggedleft\arraybackslash}X}
\begin{tabularx}{1.0\columnwidth}{L cc cc cc cc cc}
\toprule
\multirow{2}{*}{Method} & \multicolumn{2}{c}{Literal QAs} & \multicolumn{2}{c}{Inferential QAs} & \multicolumn{2}{c}{Reasoning QAs}  & \multicolumn{2}{c}{Overall}  & \multicolumn{2}{c}{Extraction} \\
\cmidrule(lr){2-3}  \cmidrule(lr){4-5}  \cmidrule(lr){6-7} \cmidrule(lr){8-9}  \cmidrule(lr){10-11}
 & R-Acc & GPT-Acc & R-Acc & GPT-Acc & R-Acc & GPT-Acc & R-Acc & GPT-Acc & F1 & RNSS \\
 \midrule
GPT-4o & \textbf{43.98} & \textbf{47.62} & \textbf{57.23} & \textbf{59.43} & \textbf{23.08} & \textbf{26.15} & \textbf{41.40} & \textbf{44.40} & \textbf{66.10} & \textbf{88.56} \\
\midrule
LLaVA$_{\text{13B}}$ & 8.96 & 8.68 & 24.91 & 21.76 & 2.46 & 2.46 & 12.11 & 10.97 & 9.22 & 45.82 \\
ChartLlama$_{\text{13B}}$ & 21.29 & 21.01 & 38.05 & 35.22 & 8.92 & 8.62 & 22.60 & 21.50 & 11.40 & 64.14 \\
ChartInstruct$_{\text{7B}}$ & 26.33 & 21.07 & 43.40 & 32.81 & 13.23 & \underline{25.23} & 27.5 & 26.15 & 8.65 & 39.73 \\
ChartAst$_{\text{13B}}$ & 27.64 & 24.09 & 38.22 & 32.70 & 14.65 & 17.97 & 26.55 & 24.87 & 16.73 & 68.04 \\
\midrule
\ours & \underline{34.45} & \underline{44.82} & \underline{56.92} & \underline{58.18} & \underline{21.85} & 21.23 & \underline{37.50} & \underline{41.40} & \underline{28.42} & \underline{75.44} \\
\bottomrule
\end{tabularx} 
\label{exp:benchmark_main}
\end{table}

We evaluate existing models on our benchmark, and the results are provided in~\cref{exp:benchmark_main}. We compare our model with four previous works, including ChartLlama~\citep{han2023chartllama}, LLaVA~\citep{liu2024visual}, ChartInstruct~\citep{masry2024chartinstruct}, and ChartAst~\citep{meng2024chartassisstant}. As shown in the table, our model consistently outperforms all existing works across three different QA levels and the raw data extraction task. We note that, compared to existing benchmarks, our benchmark includes chart images with a greater variety of chart types (i.e., 20 different chart types) and features comprehensive QAs for each image, along with raw data extraction to assess the fundamental understanding of scientific charts. Therefore, outperforming previous works demonstrates that our model has broader understanding of the chart domain and a more comprehensive understanding of the underlying data. Lastly, we also evaluate the GPT model on our benchmark. We find that while the GPT model excels at capturing global concepts~(i.e., inferential QAs), its performance on raw data understanding tasks (i.e., literal QAs and Extraction) is below $50\%$ accuracy. Moreover, for reasoning QAs, it achieves only $\approx 25\%$ accuracy, highlighting its lack of mathematical reasoning ability with chart data. We note that evaluating GPT-4o vision on the entire benchmark would be expensive. To deal with this while having fair comparison, all results in this table are evaluated on the same small subset of the benchmark, consisting of 1,000 randomly sampled chart QA pairs.

\section{Extra Experimental results}

\begin{table}[t]
\footnotesize
\centering
\caption{\textbf{Ablation of stage-1 training.}
This empirically verifies that pre-training basic chart visual perception is still important, even with abundant stage-2 instruction fine-tuning data. Moreover, learning to predict JSON data is beneficial, even on top of pre-training with descriptive captions.
}
\vspace{-0mm}
\newcolumntype{C}{>{\centering\arraybackslash}X}
\newcolumntype{L}{>{\raggedright\arraybackslash}X}
\newcolumntype{R}{>{\raggedleft\arraybackslash}X}
\begin{tabularx}{1.0\columnwidth}{L cc ccc}
\toprule
\multirow{2}{*}{Training data}  & \multicolumn{2}{c}{ChartQA} & \multicolumn{3}{c}{\ourbenchmark}  \\  \cmidrule(lr){2-3}  \cmidrule(lr){4-6} 
 & human & augmented & literal & inferential & reasoning \\
 \midrule
LLaVA-CC3M-Pretrain pairs~{\tiny\citep{liu2024visual}} & 44.80 & 83.92 & 41.45 & 34.09 & 22.31  \\
\hspace{3mm} + Chart-description pairs & 48.56 & 86.89 & 42.71 & 33.68 & 23.51 \\
\hspace{6mm} + Chart-JSON data pairs & \textbf{52.28} & \textbf{87.68} & \textbf{44.96} & \textbf{34.94} & \textbf{24.61} \\
\bottomrule
\end{tabularx}
\vspace{-3mm}
\label{exp:abltion_pretrain}
\end{table}

\begin{table}[t]
\footnotesize
\centering
\caption{\textbf{Ablation of stage-2 training.} 
Each type of new instruction / QA data improves the final performance consistently across almost all metrics.
Best result is highlighted in \textbf{Bold} and the second best is \underline{underlined}. $^\dagger$ denotes inference technique without extra data.}
\vspace{-0mm}
\newcolumntype{C}{>{\centering\arraybackslash}X}
\newcolumntype{L}{>{\raggedright\arraybackslash}X}
\newcolumntype{R}{>{\raggedleft\arraybackslash}X}
\begin{tabularx}{1.0\columnwidth}{L cc ccc}
\toprule
\multirow{2}{*}{Training data}  & \multicolumn{2}{c}{ChartQA} & \multicolumn{3}{c}{\ourbenchmark}  \\  \cmidrule(lr){2-3}  \cmidrule(lr){4-6} 
 & human & augmented & literal & inferential & reasoning \\
 \midrule
LLaVA-Instruct-150K QAs & 45.84 & 86.48 & 16.54 & 15.99 & 6.57  \\
\hspace{3mm}+ description and summary QAs & 47.04 & \textbf{87.76} & 19.90 & 15.69 & 5.26 \\
\hspace{6mm}+ Literal / infer. / reasoning QAs & 48.96 & 87.52 & 40.55 & 33.33 & 21.30 \\
\hspace{9mm}+ JSON-only QAs & 49.60 & 87.36 & 41.45 & 34.84 & 22.36 \\
\hspace{12mm}+ Data-driven QAs & \underline{52.28} & \underline{87.68} & \underline{44.96} & \underline{34.94} & \underline{24.61} \\
\cmidrule(lr){1-6}
\hspace{15mm}+ Data Prompting$^\dagger$ & \textbf{56.96} & 87.60 & \textbf{52.00} & \textbf{41.75} & \textbf{31.90} \\
\bottomrule
\end{tabularx}
\vspace{-5mm}
\label{exp:abltion_finetune}
\end{table}
\subsection{Stage 1: Pre-training for chart feature alignment}
\label{method:pretraining}

In the first training stage, the goal is to align visual and linguistic features so that visual data can be seamlessly translated into the textual domain for LLM comprehension. Employing a strategy from \citet{liu2024visual}, we use a projector to translate visual features from ViT~\citep{dosovitskiy2020image} into the textual domain, training it with pairwise image-caption data to enhance its capability to capture visual information.
We explore three configurations: utilizing only LLaVA CC3M Pretraining data,\footnote{\url{https://huggingface.co/datasets/liuhaotian/LLaVA-CC3M-Pretrain-595K}} combining LLaVA data with chart-description pairs, and using LLaVA data with both chart-description and chart-raw data pairs. The data for stage two training remains consistent across these settings, summary QAs, description QAs, three-level QAs, text-only QAs, and data-driven QAs, as depicted in \cref{fig:framework}~(b). In stage three, all models undergo LoRA fine-tuning on the downstream dataset, using LLaVA-7B as the baseline for this comparison. Results are detailed in \cref{exp:abltion_pretrain}.

\paragraph{Dense data alignment is beneficial for both chart data comprehension and reasoning.}
For chart images, chart-description pairs act as standard image-caption pairs. However, to more effectively bridge the visual-textual gap, we also utilize chart-json pairs that encompass the underlying numerical data and its schema of the charts. This approach not only aligns visual features with textual descriptions but also significantly enhances model performance, as demonstrated by improvements of approximately 2\% in literal QAs and about 1\% in reasoning skills, according to results in \cref{exp:abltion_pretrain}.

\subsection{Stage 2: End-to-end fine-tuning}
\label{method:finetuning}

The second stage, end-to-end fine-tuning, trains the MLLM to actually understand the aligned visual tokens so that it follows the user instruction and reason about the answer, on top of the inherent language capability from the original LLM.
We utilize a significant number of image-QA pairs to jointly tune the LLM and the projector.
To evaluate the effectiveness of incorporating chart QAs during fine-tuning, we conduct ablation studies starting with a baseline that uses only LLaVA Instruct-150K data,\footnote{\url{https://huggingface.co/datasets/liuhaotian/LLaVA-Instruct-150K}} incrementally adding extra QA pairs.
All methods leverage the same pre-training weights, derived from training on LLaVA data with both chart-description and chart-raw data pairs~(the best setting in~\cref{method:pretraining}).
In stage three, all models undergo LoRA fine-tuning on the downstream dataset. Comprehensive results are presented in \Cref{exp:abltion_finetune}.

\paragraph{JSON-only QAs allow transferring pure text reasoning abilities to multimodal chart understanding.}
The chart understanding of MLLMs can be seen as two stages: visual and text raw data alignment~(which is done in the training of the first stage) and question answering with reasoning ability on the raw textual data~(JSON).
Thus, with a well-aligned first stage training, we hypothesize that re-blending some pure textual QAs, preserving the ability of reasoning on text raw data, can also benefit the reasoning abilities in visual-text scenarios.
As detailed in~\cref{method:qa}, for JSON-only QAs, rather than utilizing chart images and QAs, we replace the chart image with JSON data and a README, resulting in purely text-based QAs for training.
\Cref{exp:abltion_finetune} demonstrates the effectiveness of each QA type.
We discover that re-blending JSON-only data during the end-to-end fine-tuning stage improves chart reasoning skills, matching the assumption.

\paragraph{Data-driven QAs in the fine-tuning stage enable MLLMs to enhance prediction accuracy through data prompting.}
As detailed in~\cref{method:qa}, data-driven QAs are multi-turn QAs, which require models to extract raw data before answering given questions. Combined with the raw data reasoning abilities enhanced via JSON-only QAs, the model can perform data prompting during inference, where models achieve better reasoning robustness by first extracting raw data and then answering the given question based on the data. Please refer to~\cref{suppl:dataset:demo_cor} for some examples.
As shown in~\cref{exp:abltion_finetune}, data-driven QAs significantly enhance the model's ability to capture visual information.
Furthermore, leveraging data prompting in inference significantly improves performance across all downstream tasks.

\clearpage
\section{Examples for JSON template and READMEs}
\label{suppl:dataset:template}
\subsection{Example 1}

\begin{figure*}[h]
  \centering
  \includegraphics[page=6, trim={0 0 0 0}, clip, width=1.\textwidth]{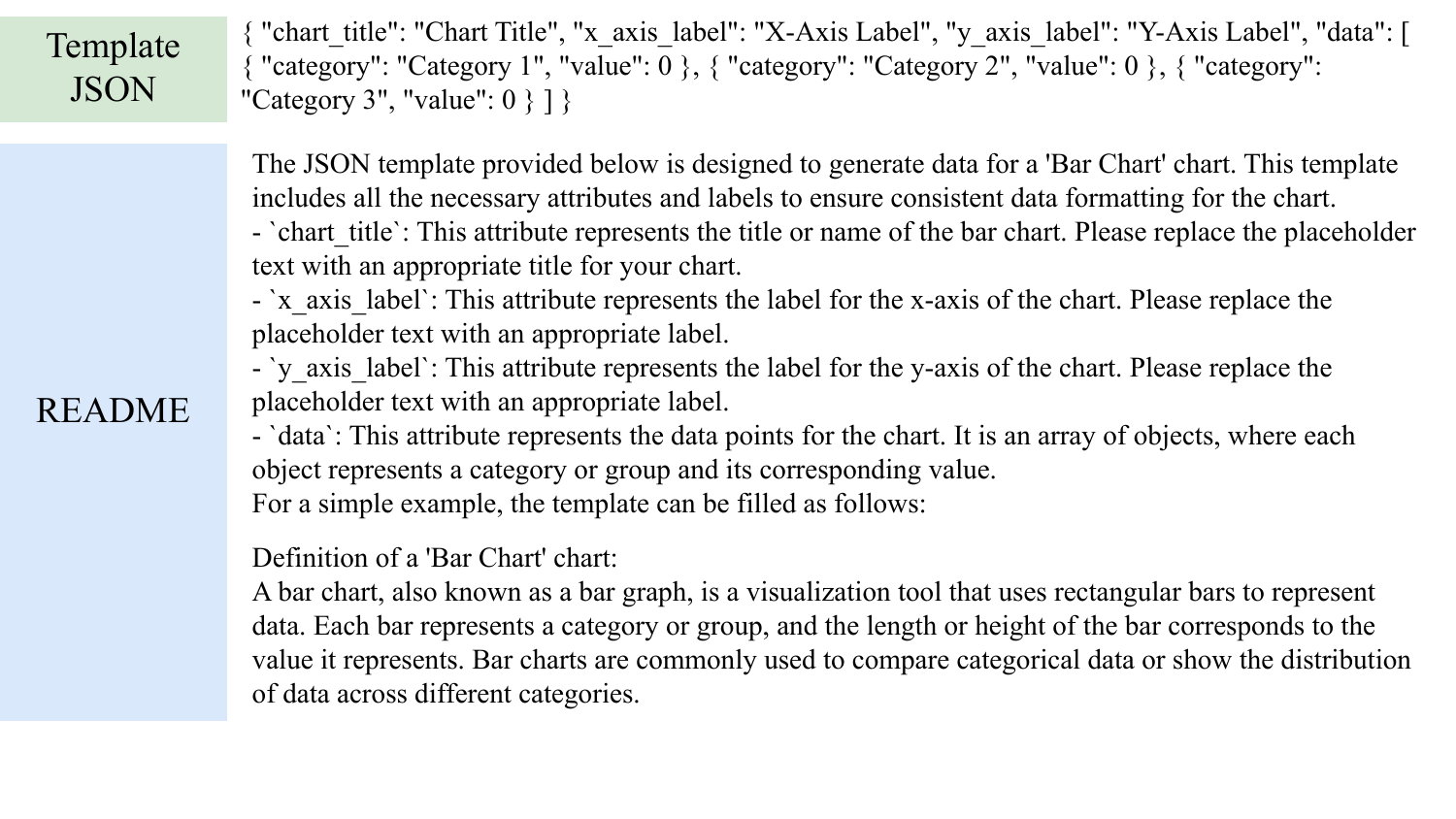}
  \vspace{-5mm}
\end{figure*}

\subsection{Example 2}

\begin{figure*}[h]
  \centering
  \includegraphics[page=7, trim={0 0 0 0}, clip, width=1.\textwidth]{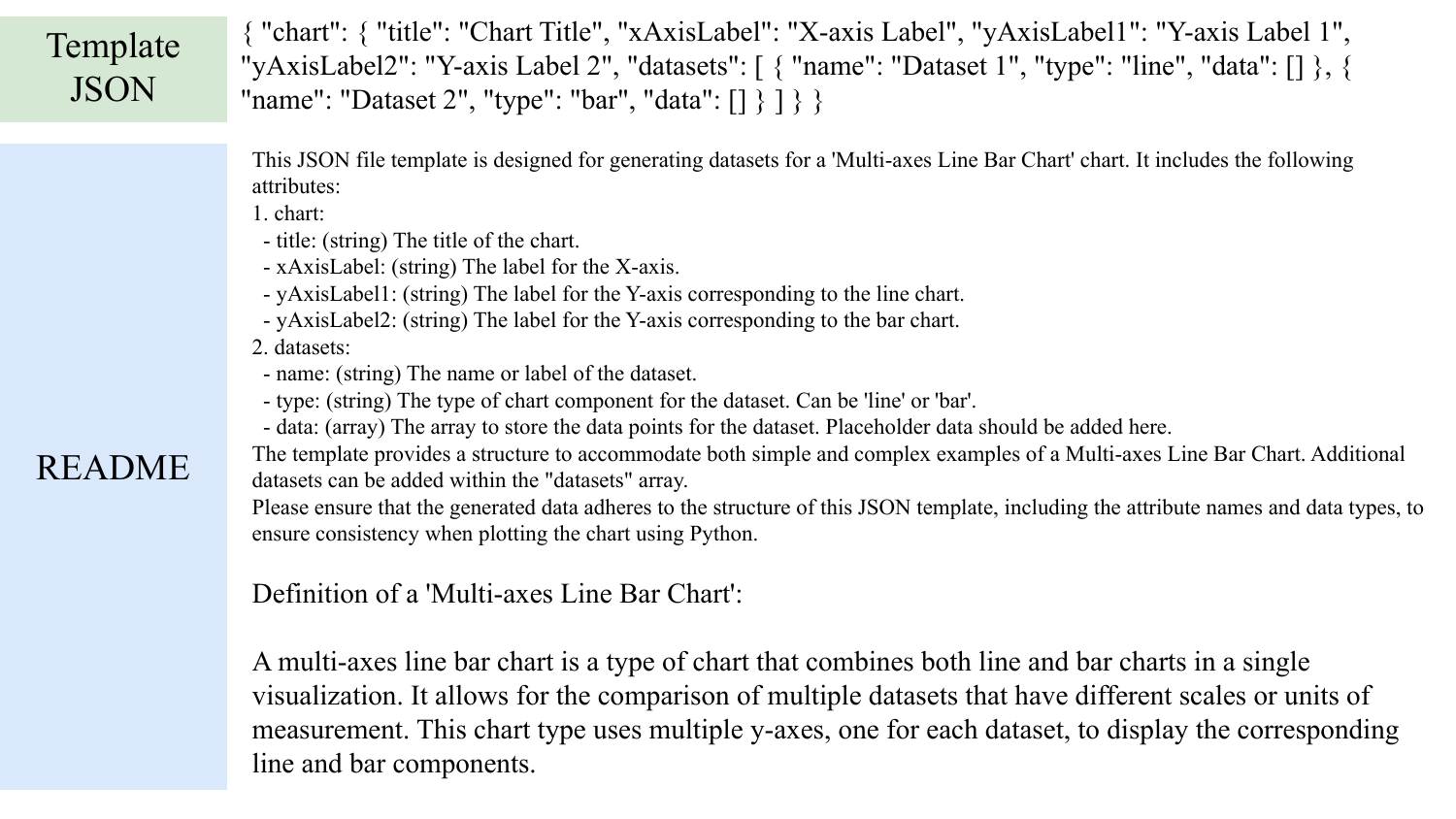}
  \vspace{-5mm}
\end{figure*}

\clearpage

\section{Examples for pre-defined topics}
\label{suppl:dataset:topics}

\begin{figure*}[h]
  \centering
  \includegraphics[page=9, trim={0 0 0 0}, clip, width=1.\textwidth]{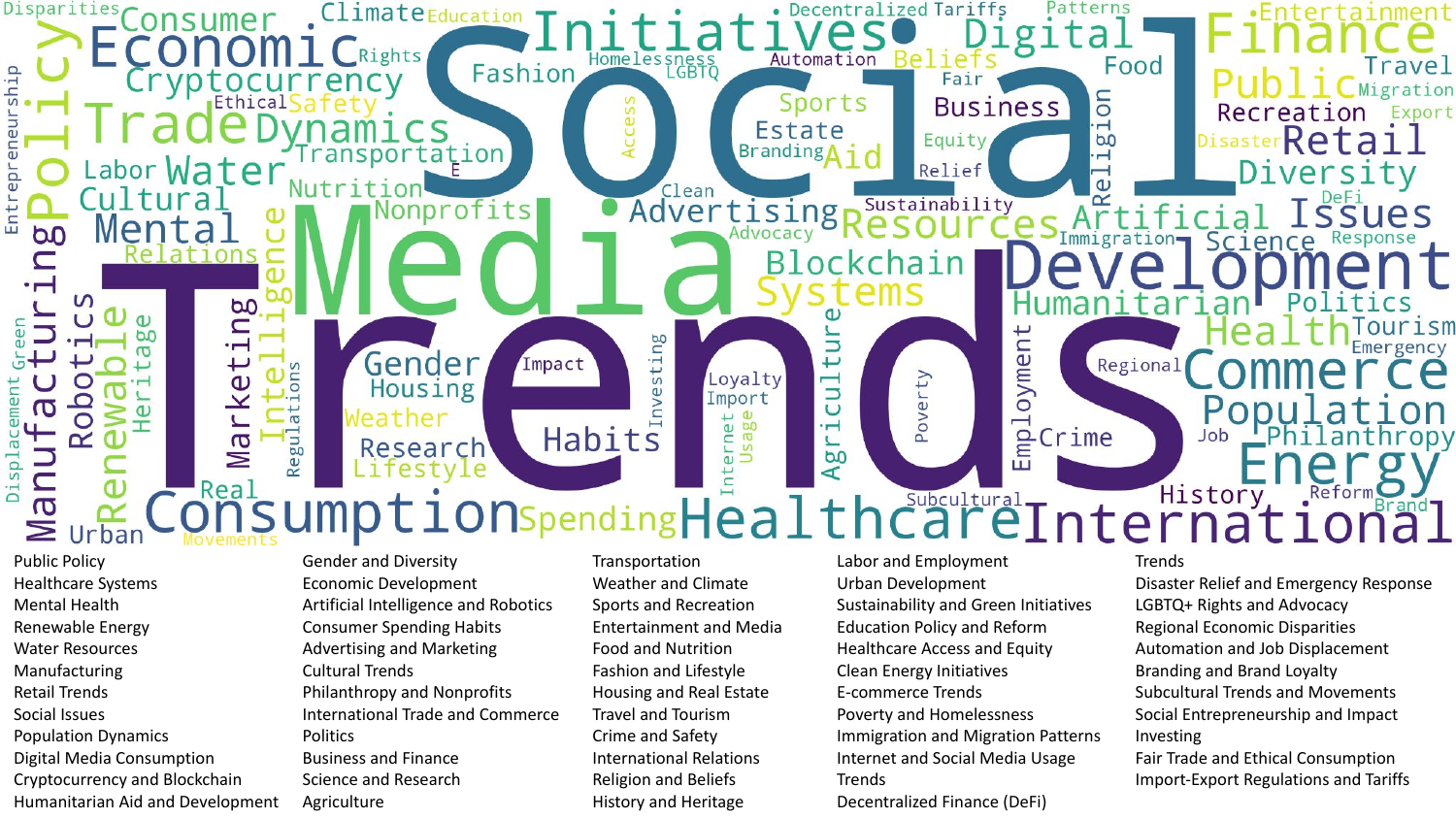}
  \vspace{-5mm}
\end{figure*}

We provide a word cloud in the figure above to show the frequency of each word in the defined topic set. A comprehensive list of all the topics is also provided at the bottom of this figure.
\clearpage

\section{Examples of augmented QAs}
\label{suppl:dataset:demo_qa}
\subsection{JSON-only QA: example 1}

\begin{figure*}[h]
  \centering
  \includegraphics[page=10, trim={0 0 0 0}, clip, width=1.\textwidth]{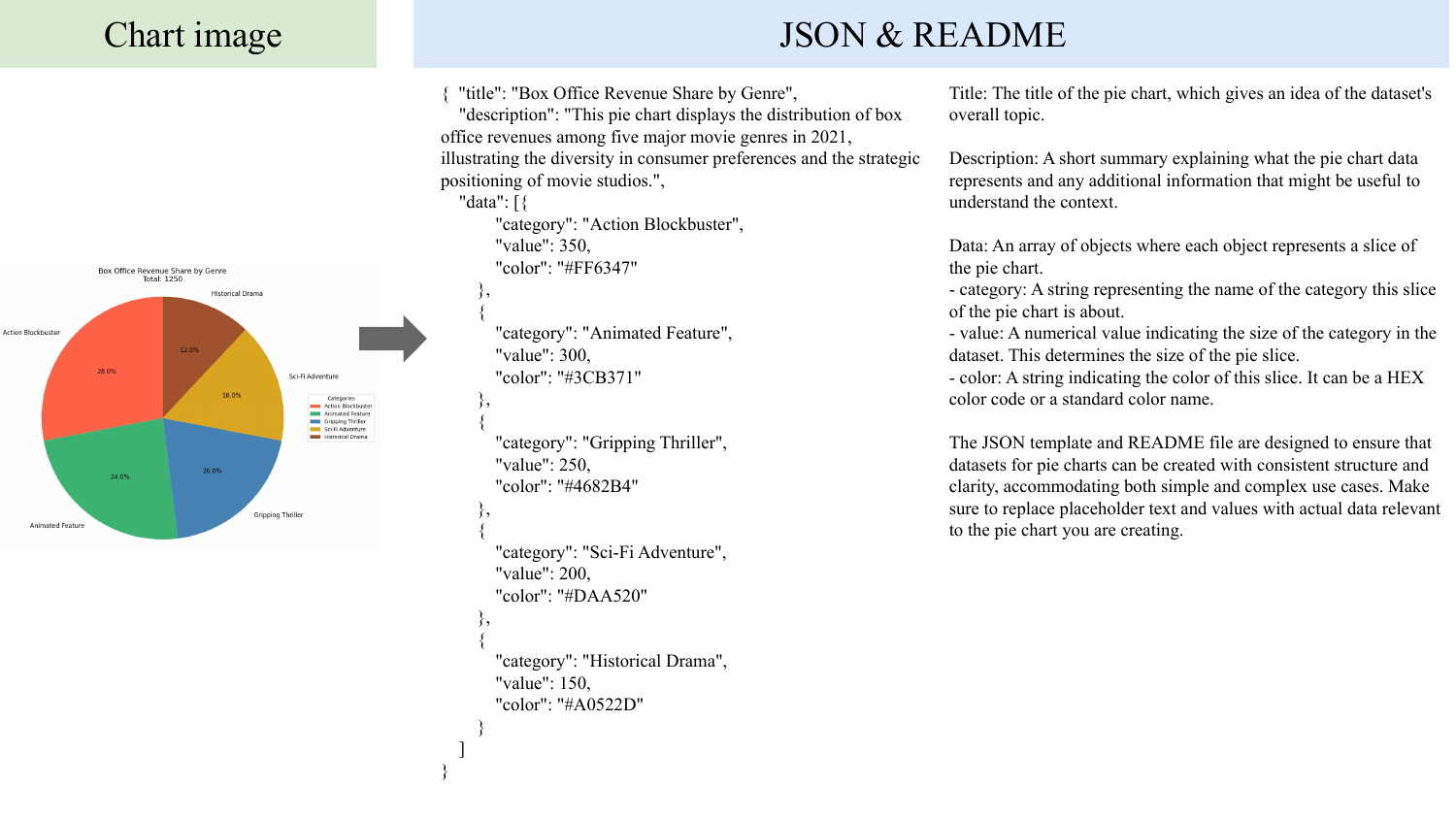}
  \vspace{-5mm}
\end{figure*}

\subsection{JSON-only QA: example 2}

\begin{figure*}[h]
  \centering
  \includegraphics[page=11, trim={0 0 0 0}, clip, width=1.\textwidth]{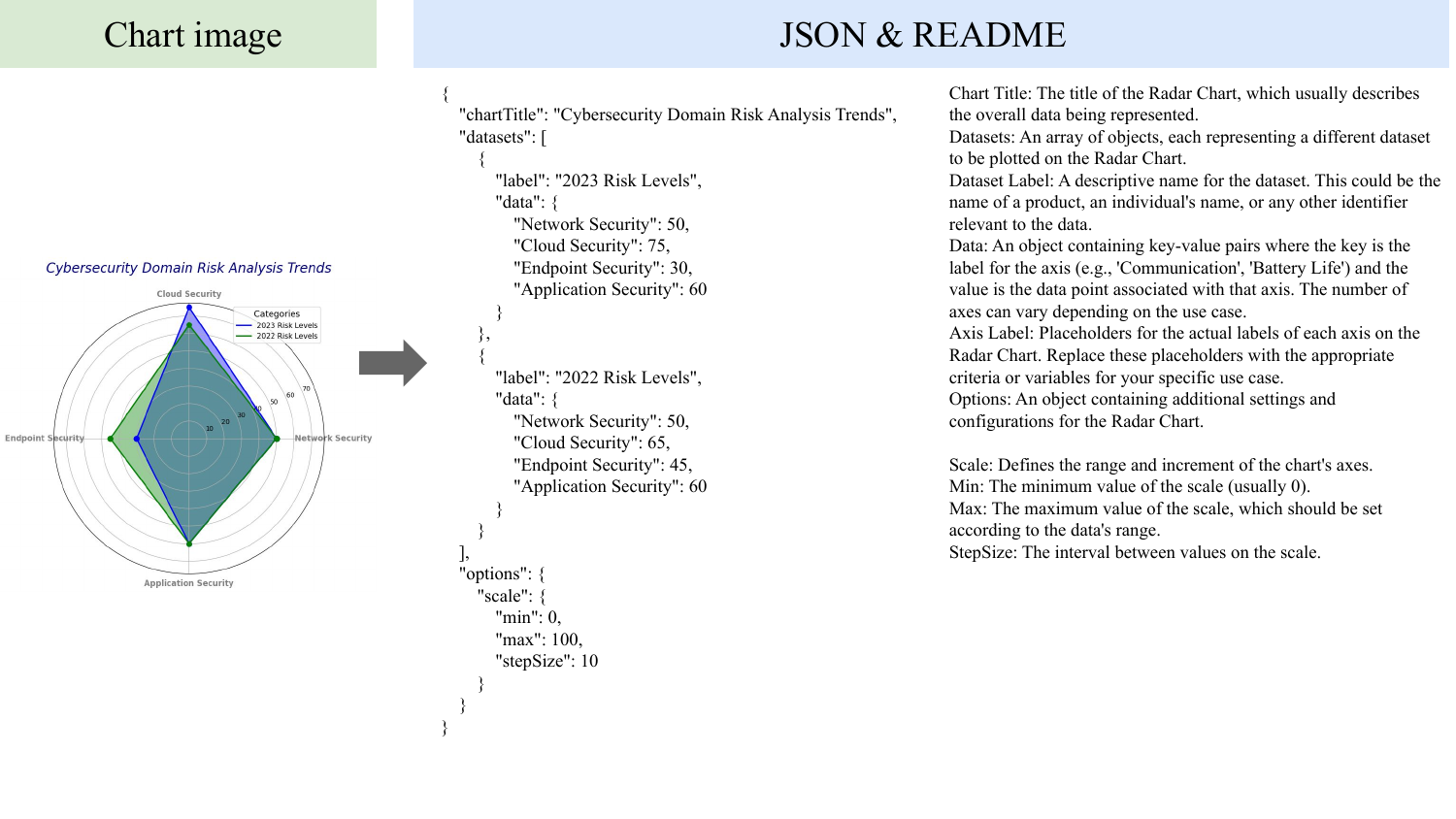}
  \vspace{-5mm}
\end{figure*}

\clearpage
\subsection{Data-driven QA: example 1}

\begin{figure*}[h]
  \centering
  \includegraphics[page=12, trim={0 0 0 0}, clip, width=1.\textwidth]{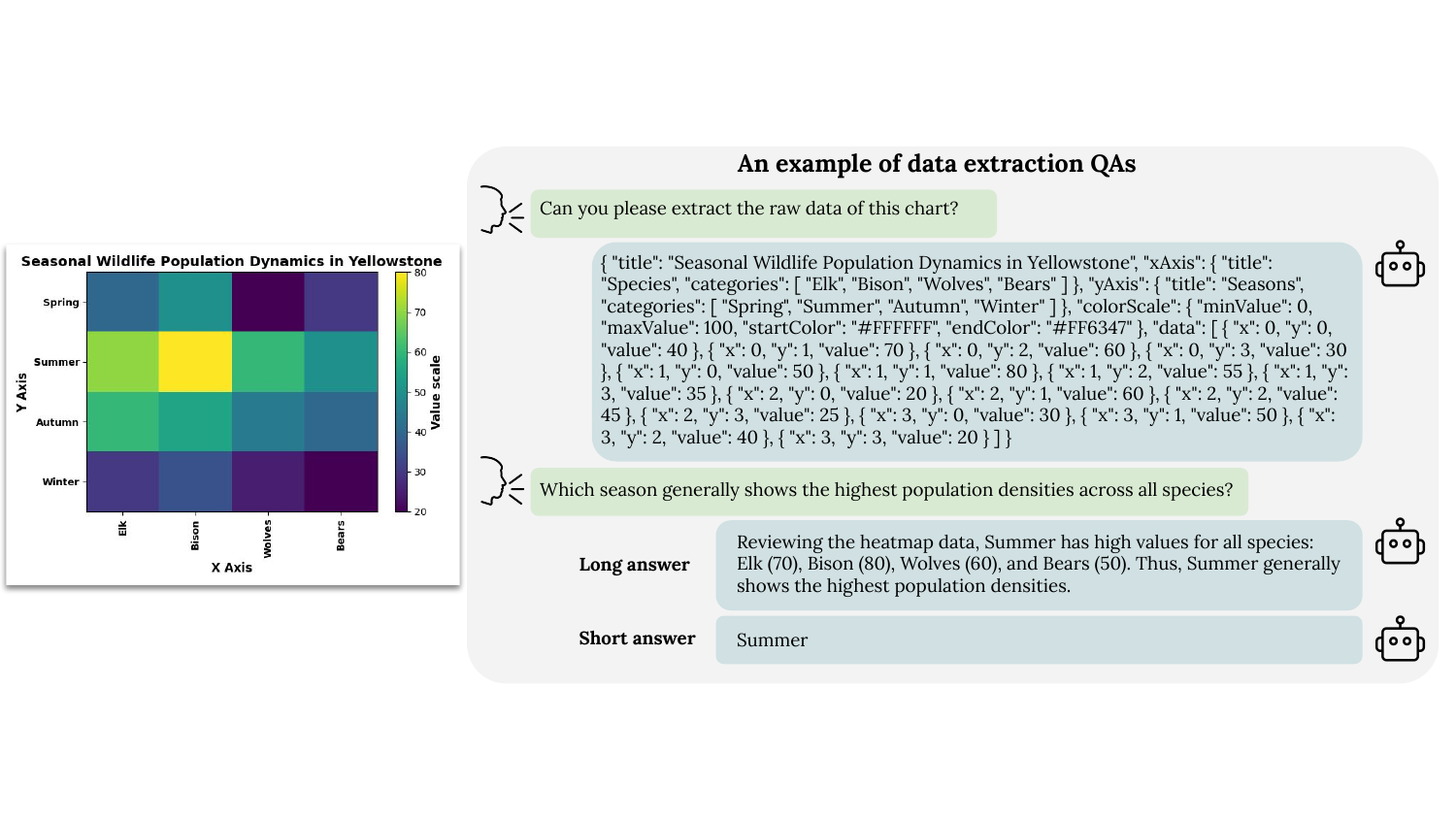}
  \vspace{-5mm}
\end{figure*}

\subsection{Data-driven QA: example 2}

\begin{figure*}[h]
  \centering
  \includegraphics[page=13, trim={0 0 0 0}, clip, width=1.\textwidth]{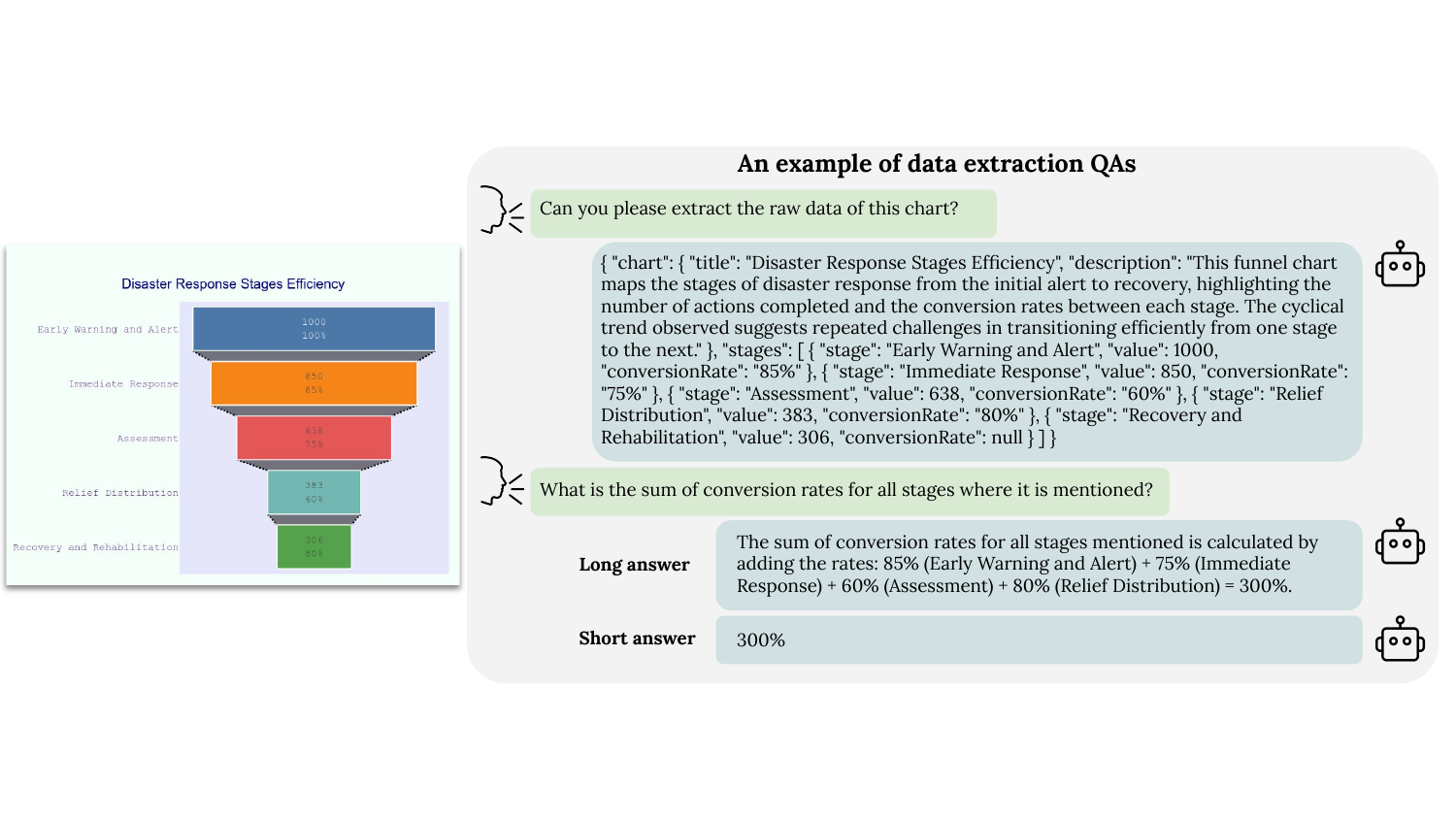}
  \vspace{-5mm}
\end{figure*}

\clearpage

\section{Examples of data prompting}
\label{suppl:dataset:demo_cor}
\subsection{Example 1}

\begin{figure*}[h]
  \centering
  \includegraphics[page=14, trim={0 0 0 0}, clip, width=1.\textwidth]{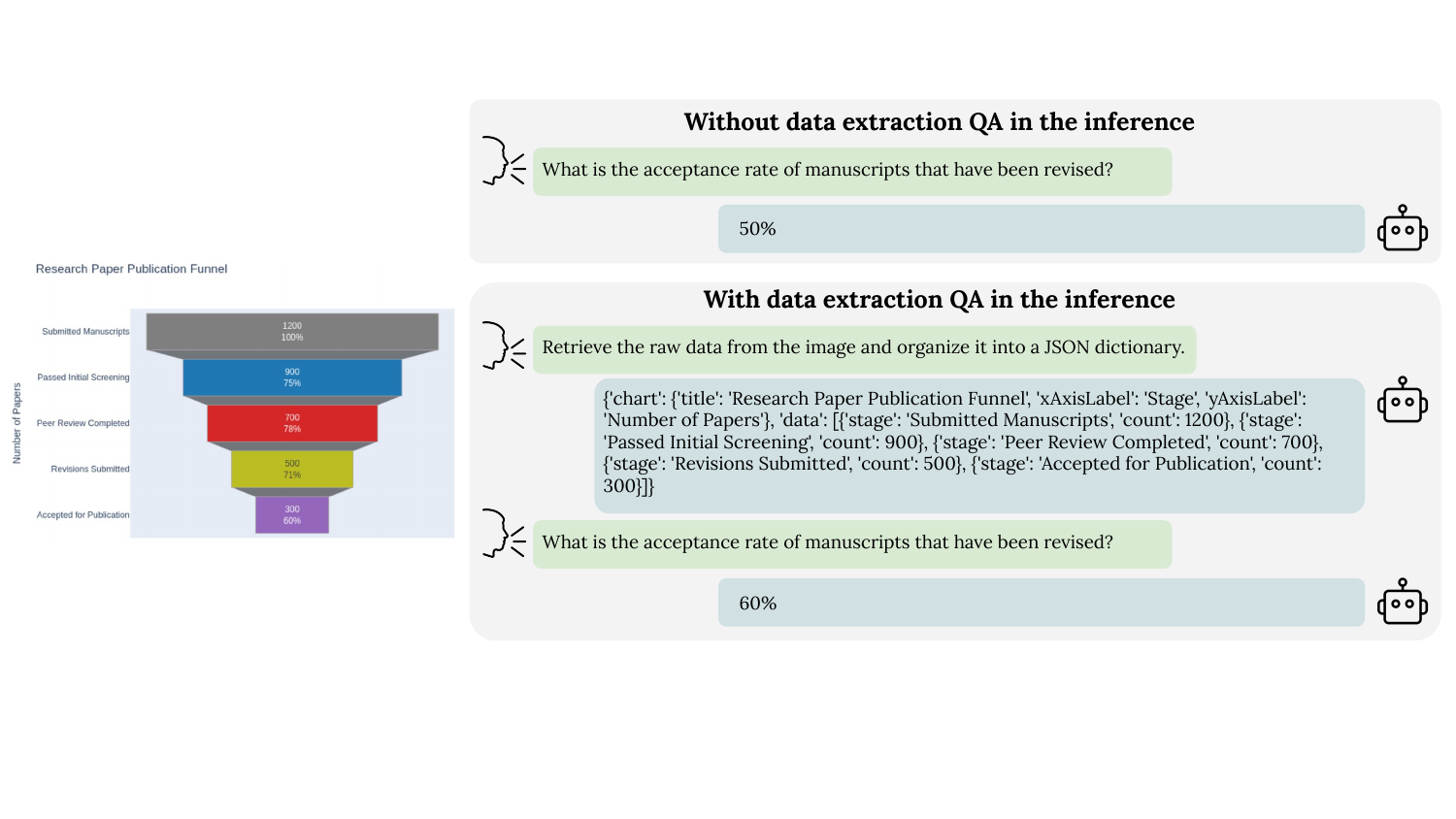}
  \vspace{-5mm}
\end{figure*}

\subsection{Example 2}

\begin{figure*}[h]
  \centering
  \includegraphics[page=15, trim={0 0 0 0}, clip, width=1.\textwidth]{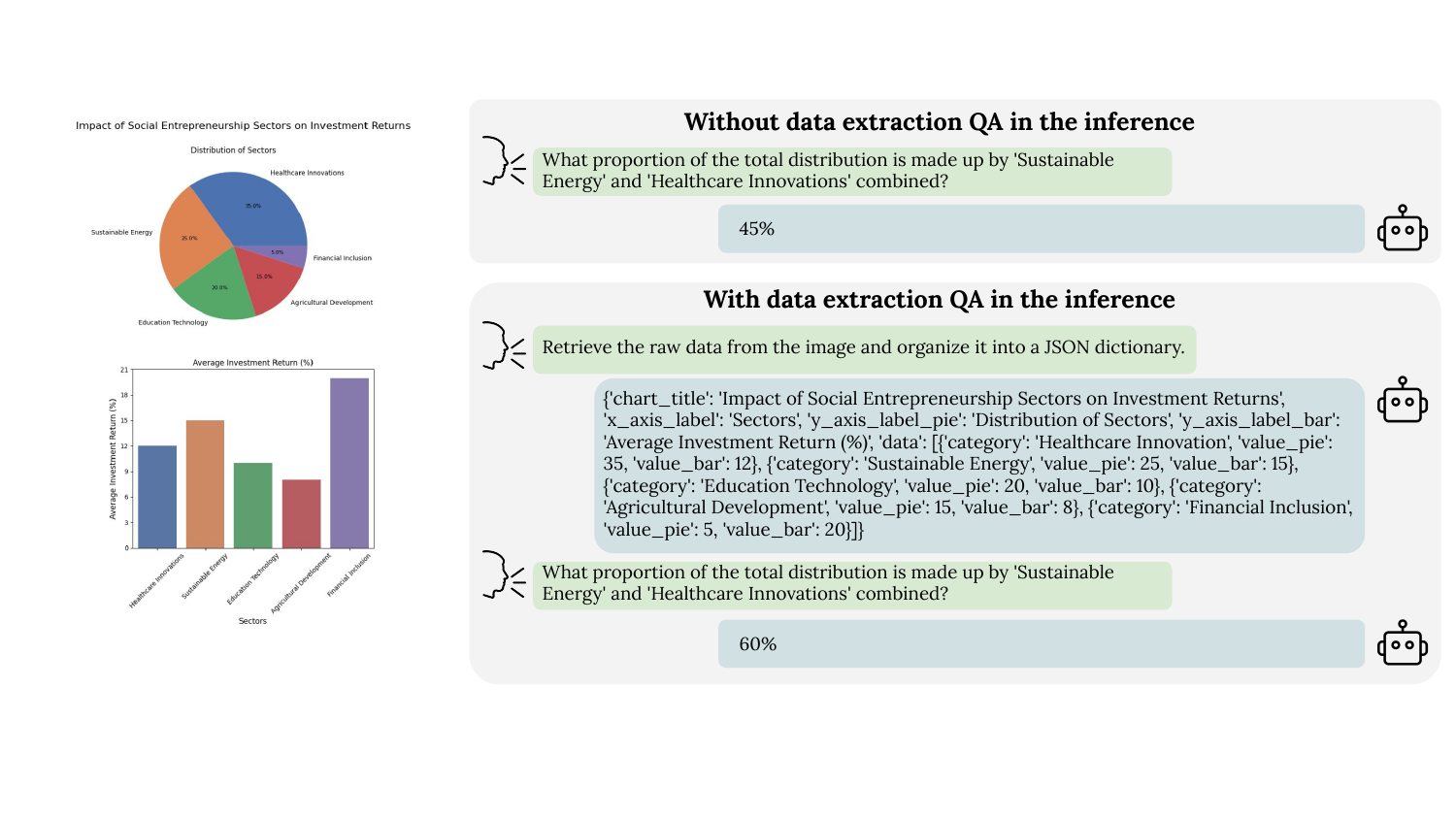}
  \vspace{-5mm}
\end{figure*}

\clearpage

\section{Examples from our benchmark}
\label{suppl:dataset:demo}
\subsection{Example 1}

\begin{figure*}[h]
  \centering
  \includegraphics[page=1, trim={20 40 20 20}, clip, width=1.\textwidth]{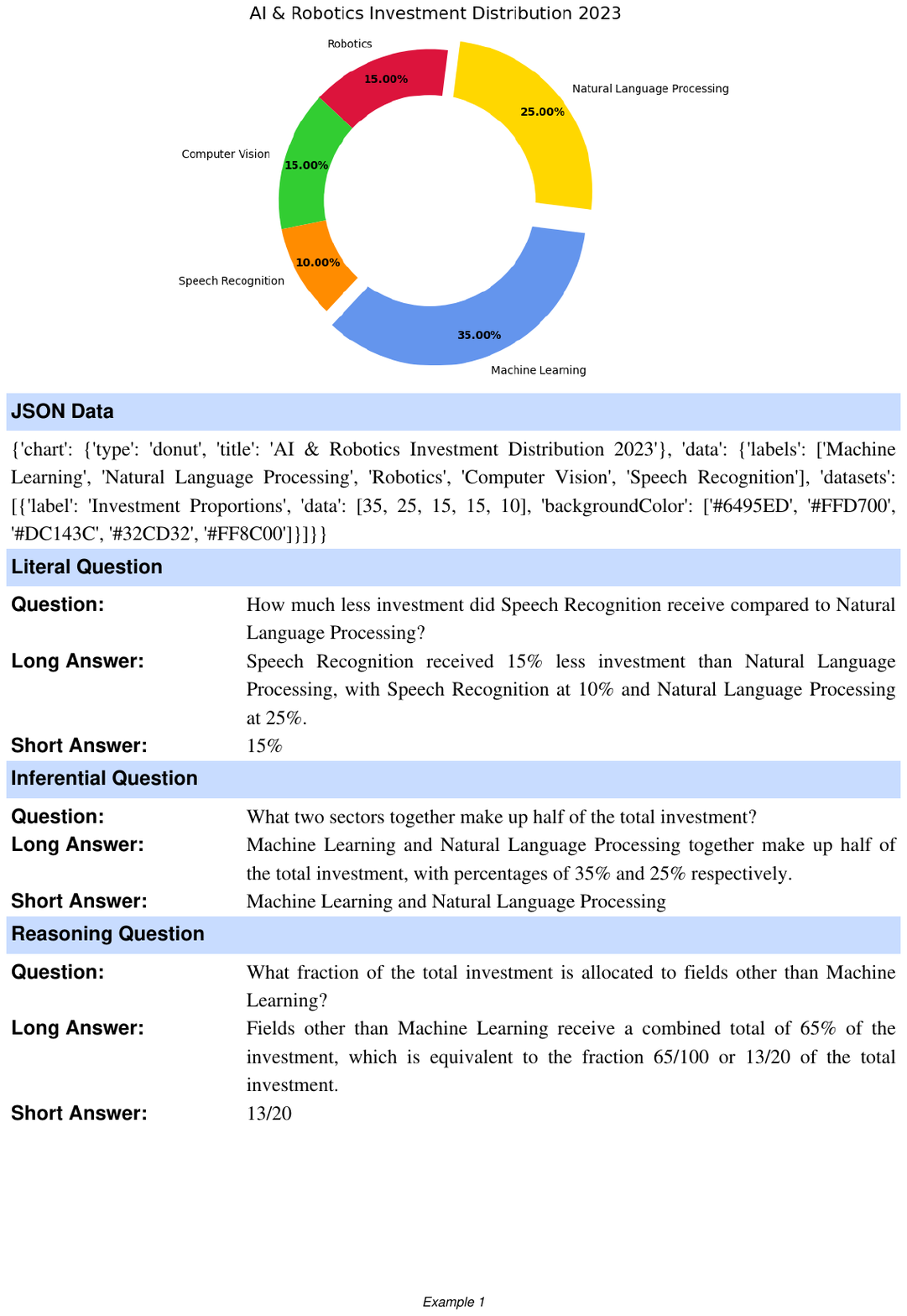}
  \vspace{-5mm}
\end{figure*}

\clearpage
\subsection{Example 2}

\begin{figure*}[h]
  \centering
  \includegraphics[page=2, trim={20 40 20 20}, clip, width=1.\textwidth]{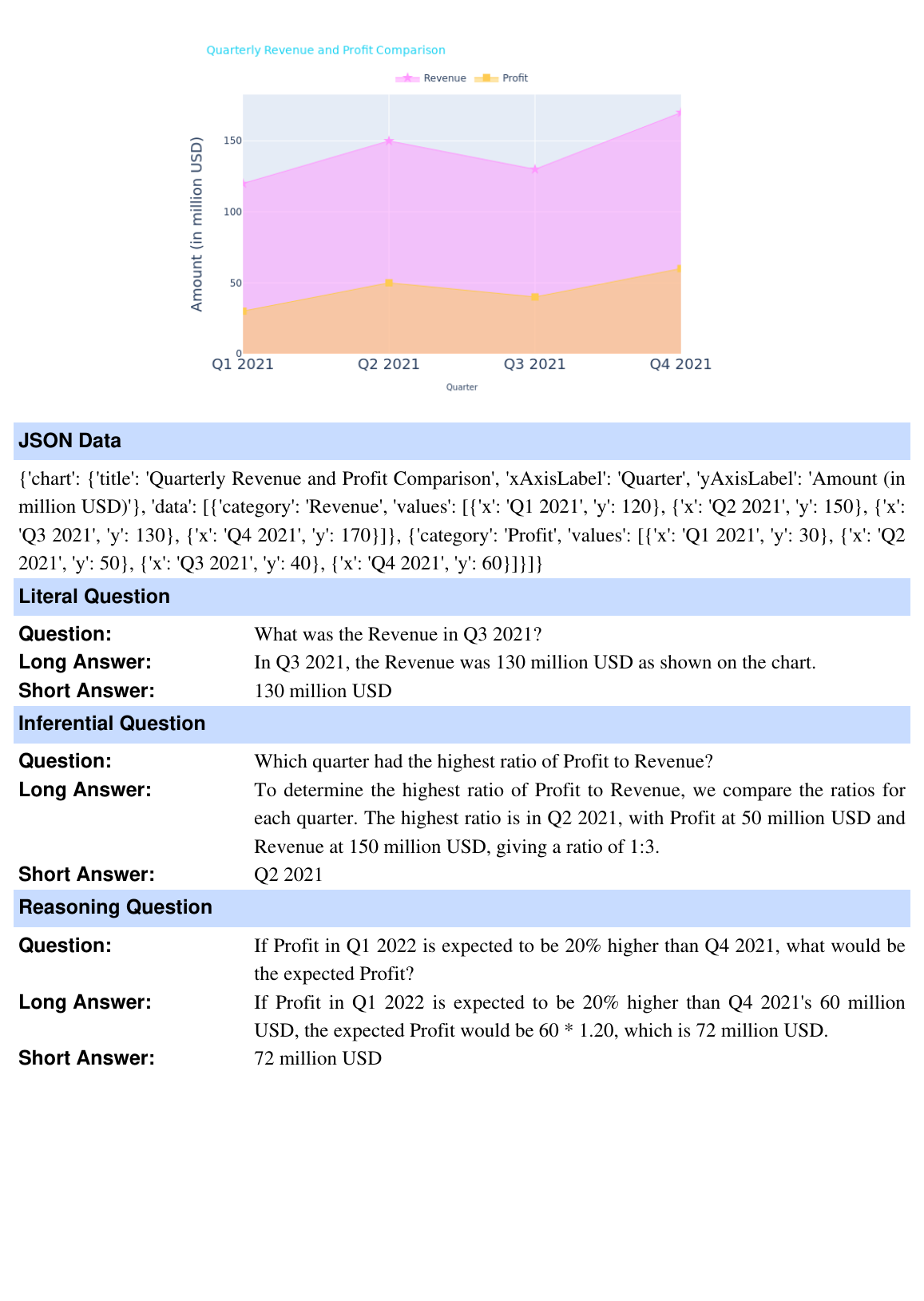}
  \vspace{-5mm}
\end{figure*}

\clearpage
\subsection{Example 3}

\begin{figure*}[h]
  \centering
  \includegraphics[page=3, trim={20 40 20 20}, clip, width=1.\textwidth]{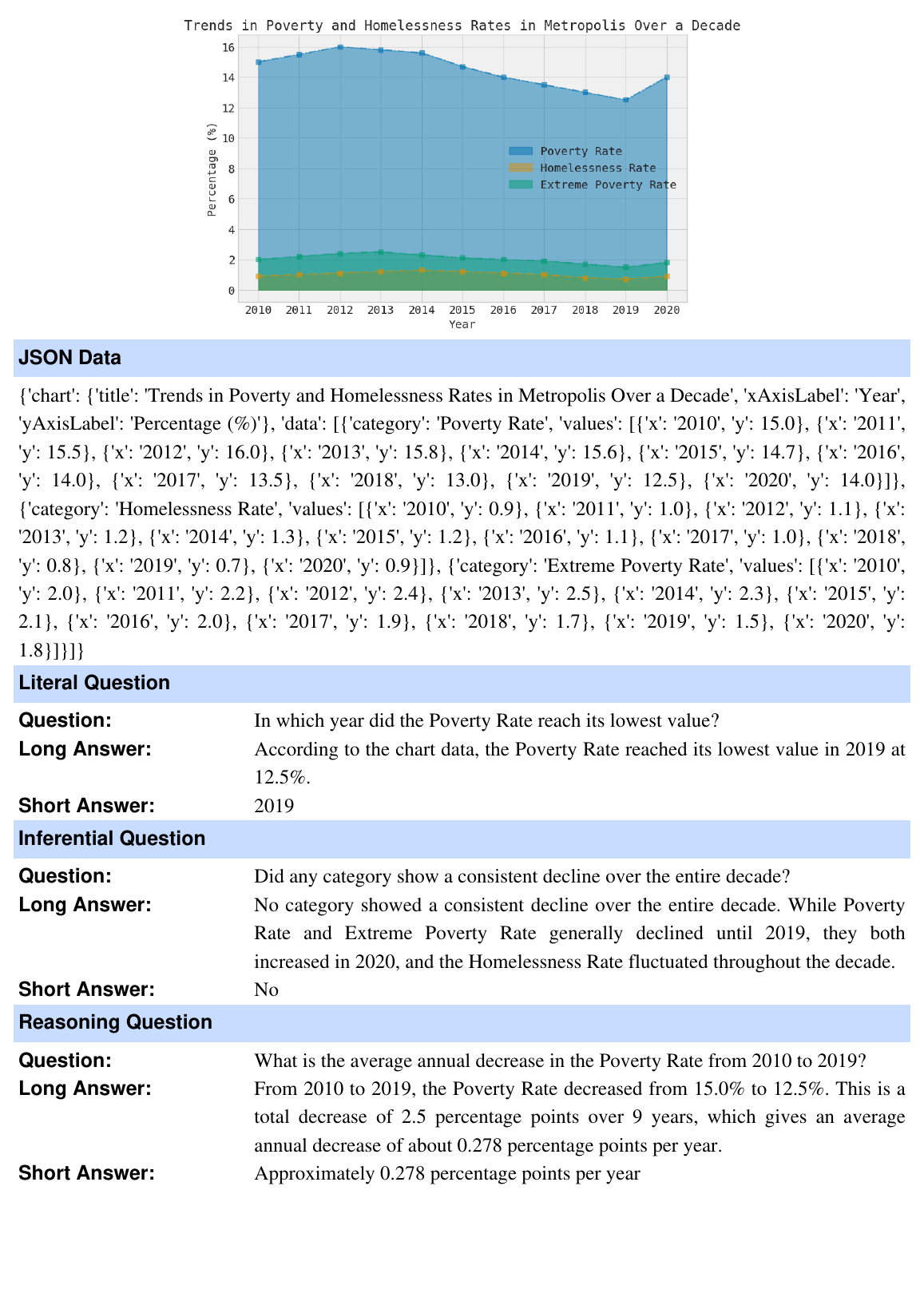}
  \vspace{-5mm}
\end{figure*}

\end{document}